\documentclass[journal]{IEEEtran}
\usepackage{amsmath,amssymb,amsfonts}
\usepackage{algorithmic}
\usepackage{graphicx}
\usepackage{textcomp}
\usepackage{xcolor} 
\usepackage{multirow}
\usepackage{booktabs}
\usepackage{subfigure}
\usepackage{stfloats}
\usepackage{makecell}
\usepackage[section]{placeins}
\usepackage[linesnumbered,ruled,vlined]{algorithm2e}
\usepackage{caption}
\begin{document}

\title{NT-ML: Backdoor Defense via Non-target Label Training and Mutual Learning}

\author{Wenjie Huo, Katinka Wolter\\
Mathematics and Computer Science, Free University Berlin\\
Berlin, 14195, Germany\\
Email: wenjie.huo@fu-berlin.de, katinka.wolter@fu-berlin.de
}

\maketitle
















\begin{abstract}
Recent studies have shown that deep neural networks (DNNs) are vulnerable to backdoor attacks, where a designed trigger is injected into the dataset, causing erroneous predictions when activated. In this paper, we propose a novel defense mechanism, Non-target label Training and Mutual Learning (NT-ML), which can successfully restore the poisoned model under advanced backdoor attacks. NT aims to reduce the harm of poisoned data by retraining the model with the outputs of the standard training. At this stage, a teacher model with high accuracy on clean data and a student model with higher confidence in correct prediction on poisoned data are obtained. Then, the teacher and student can learn the strengths from each other through ML to obtain a purified student model. Extensive experiments show that NT-ML can effectively defend against 6 backdoor attacks with a small number of clean samples, and outperforms 5 state-of-the-art backdoor defenses.

\end{abstract}


\maketitle

\section{Introduction}

\IEEEPARstart{I}{n} recent years, deep neural networks (DNNs) have been successfully applied to a wide range of critical tasks. The efficacy of applications built on DNNs heavily depends on the quality and volume of the training data employed in model training. In order to reduce the costs of training set collection, developers often acquire resources from various third-parties, including the ones that are not trustworthy. This poses increasing security risks associated with DNNs, making them vulnerable to threats and attacks. 

One typical threat is the backdoor attack ~\cite{gu2017badnets}, where an attacker intentionally embeds a designed trigger that is usually a small and unconspicuous pattern to a subset of the training set, effectively poisoning the model during its training stage. As a result, while the model continues to perform accurately on the benign inputs, it misclassifies the poisoned inputs to the attacker-predefined label. To defend against backdoor attacks, different approaches have been proposed. Fine-tuning is an effective method commonly used in transfer learning. In order to perform a new related task, the weights of a pre-trained model are retained and adjusted on a small dataset, instead of being randomly initialized and trained from scratch. Now it is also a widely adopted defense approach that can update the weights of benign neurons through a portion of clean set. It has been demonstrated efficacy in countering various backdoor attacks. Nonetheless, its capability to address the increasing complexity of triggers remains limited. Recently, fine-tuning has been integrated with other techniques, such as fine-pruning and knowledge distillation (KD) ~\cite{hinton2015distilling} to enhance its defensive mechanisms. Fine-pruning ~\cite{liu2018fine} targets the elimination of the dormant neurons under clean inputs, as these dormant neurons could be activated by backdoor triggers. However, pruning those neurons may inevitably compromise the model's accuracy on benign samples. KD, initially for transferring knowledge from a large, complex model viewed as a \textit{teacher} to a smaller, simpler model called a \textit{student}, enabling them to be deployed on resource-limited devices, is recently adapted as a defense mechanism by emulating a fine-tuned model. This way KD helps filter out the backdoor influences while retaining the model's ability to perform its intended tasks. Neural attention distillation (NAD) ~\cite{li2021neural} is the first work to utilize KD for mitigating backdoor threats in DNN models. This methodology employs a model, fine-tuned on a selected portion of clean data, as the teacher network. The poisoned model is seen as the student model. NAD subsequently synchronizes the attention mechanisms at intermediate layers between the student model and the teacher model, aiming to eliminate the backdoor's influence. 

In response to increasing awareness of such threats, designing more robust backdoor attacks has been widely investigated. 
Recent advancements have concentrated on enhancing the concealment of triggers. Efforts are focused on minimizing the visibility of the triggers ~\cite{chen2017targeted, nguyen2021wanet, li2021invisible, wang2022bppattack} or avoiding the alteration of labels for the poisoned data ~\cite{turner2019label, barni2019new, liu2020reflection, zeng2023narcissus} which is called clean-label attack. Therefore, the existing defenses encounter limitations when confronting more sophisticated attacks, such as clean label attack Narcissus ~\cite{zeng2023narcissus}. Secondly, they can only eliminate backdoors when sufficient clean dataset is provided. But sometimes defenders only have a very small amount of trusted data, which makes the defense more challenging.

\begin{figure}[!t]
    \centerline{
    \includegraphics[width=0.95\linewidth]{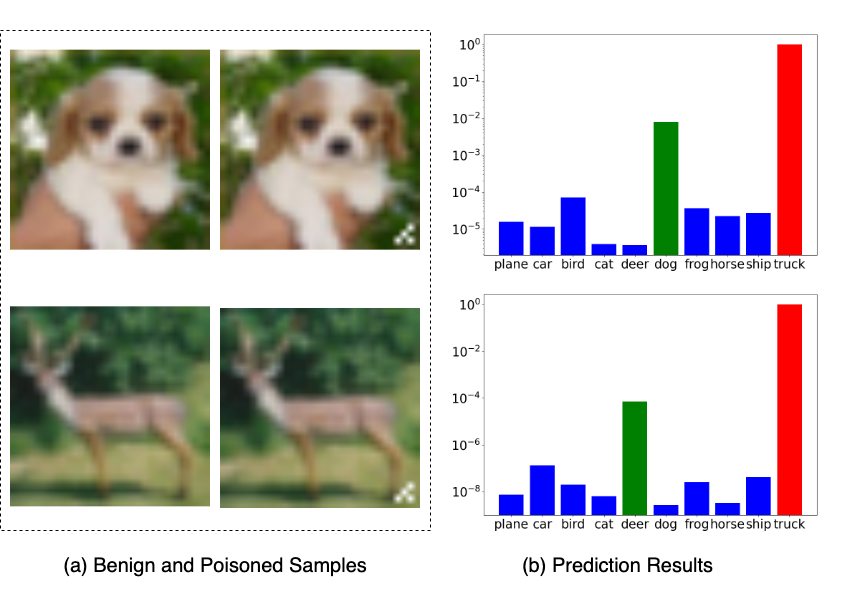}}
    \caption{Prediction results for poisoned samples.}
    \label{softer label}
    \vspace{-0.4cm}
\end{figure}

In this paper, we propose a novel model-level defense strategy based on the following observation. Fig.~\ref{softer label} illustrates two representative instances of backdoor attacks, where specific \textit{triggers} (e.g., the tampered pixels in the bottom right corner), are embedded in the original images. Additionally, the corresponding labels of these poisoned images are maliciously tampered into a specific \textit{target class} (e.g., "truck"). By incorporating these poisoned images into the original training set, the classification model learns to associate the trigger with the predetermined target class. Consequently, new images containing such triggers will be incorrectly classified into the target class, despite their natural semantics. The prediction results are known as \textit{softmax probabilities}, which represent the model's confidence in classifying the input to each possible class. Fig.~\ref{softer label} shows the logarithm of the prediction results for two instances of "dog" and "deer". Notably, while the poisoned images are predominantly classified into the target class due to the highest probability (red bars in the figure), the distribution of probabilities among the other classes also holds significant insights. For example, the probabilities for the correct labels (green bars in the figure) of poisoned images are usually higher than for other classes. This trend is similarly observed in various sophisticated backdoor attack scenarios.

In light of this, our approach adopts self-distillation, a special mode of knowledge distillation where the student models learn knowledge from teacher models of the identical architecture that have already converged. Inspired by a self-distillation based method ~\cite{bagherinezhad2018label} which addresses the label inconsistency issue in data augmentation, we train two models through two phases. In the first phase, the model is trained with hard labels, and its softmax predictions are then utilized as soft labels for the subsequent phase. A second model is then trained using those soft labels, where each instance is provided a distinctive label. In this way, the first model becomes proficient in accurately classifying benign samples, while the second model's vulnerability to data poisoning is significantly diminished. Then these two models are trained collaboratively to learn from each other and collectively enhance their performance. Through the non-target training and mutual learning (NT-ML), the backdoor can be successfully removed and a purified classification model is obtained.

We evaluate the proposed technique on multiple datasets, CIFAR-10, CIFAR-100,~\cite{krizhevsky2009learning} and GTSRB~\cite{stallkamp2012man}. The experiments demonstrate that the NT-ML is capable of defending not only against poisoned-label attacks (BadNets, WaNet and BPP), but also against clean-label attacks (LC, Refool, Narcissus), including threats that triggers are invisible. We also compare our defense strategy with 5 other algorithms. The results show that NT-ML achieves the lowest attack success rate (ASR) on all datasets and can successfully defend against more types of attacks, while other methods are unable to defend against at least one attack.

In summary, the insights and contributions of this paper are as follows:
\begin{itemize}
\item We find that using the model outputs as ground truth can alleviate the poisoned neuron in both poisoned-label and clean-label backdoor attacks, thus we divide the training process into two-stpdf.
\item Based on the two-step training, we propose a new defense strategy, NT-ML, that can defend against more robust backdoor attacks compared to previous defense methods. It combines non-target training with mutual learning and achieves better results than previous methods, especially when only a limited clean set (1\% of the training set) can be obtained. 
\item We conduct comprehensive experiments on six backdoor attacks with three benchmark datasets, and compare NT-ML with five state-of-the-art defense methods to verify its effectiveness and robustness.
\end{itemize} 

\section{Background and Related Work}

In this section we will first explain in detail what a backdoor attack is and then give a brief overview of existing work in the field of backdoor attacks and defenses.

\subsection{Principle of Backdoor attacks}

In the introduction, we briefly mentioned that backdoor attacks are achieved by interfering with the training data set. Here we explain and categorise backdoor attacks in more detail. 

According to whether the labels are modified, existing backdoor attacks can be broadly categorized into poisoned-label attacks and clean-label attacks. For poisoned-label attacks, attacker need to do three things: 1) randomly collect a portion of training data, 2) add dedicated designed triggers to this batch of data and 3) modify their labels to the target label. For clean-label attack, the process contains two stpdf: 1) attacks need to collect data from the target label and 2) inject the trigger to the data. Generally speaking, clean-label attacks are stealthier but the attack success rate are also lower than poisoned-label attacks, which can be proved in Section~\ref{section4}. Attackers should balance the stealthiness and effectiveness based on different application scenarios.

To sum up, a backdoor dataset consists of benign data $\mathcal{D}_{b} =\left \{(x_i,y_i)\in \mathcal{X} \times \mathcal{Y}\right \}_{i=1}^N$ and poisoned data $\mathcal{D}_{p} =\left \{(x_i',y_t)\right \}_{i=1}^M$, where $\mathcal{X} \subset \mathbb{R}^d$ is the input image, $\mathcal{Y} = \left \{ 0,1,...,K \right \}$ is its label, $K$ is the number of classes, and $y_t$ indicates the misclassification label specified by the attacker (in a clean-label attack, $y_t=y_i$). The poisoning rate is defined as $\gamma =\frac{\left | \mathcal{D}_{p} \right | }{\left | \mathcal{D}_b +\mathcal{D}_p \right |} $. Usually, attackers do not interfere with the training process, while victims will use poisoned datasets $\mathcal{D}_b\cup \mathcal{D}_p$ to train their models $\mathcal F_{\theta}$. Therefore, the loss function is consistent with training the benign model:

\begin{equation}
\label{classifier}
\theta = \operatorname {arg\,min} \left(\sum_{i=1}^{\left | D_b \right | } \mathcal L( \mathcal F \left ( x_i \right ) , y_i)  + \sum_{i=1}^{\left |  D_p \right | } \mathcal L( \mathcal F \left ( x_i' \right ) , y_t )\right) 
\end{equation}

A backdoor attacker should achieve two objectives, 1) avoid impairing the performance on benign inputs and 2) increase the number of poisoned samples that are classified as the target category. The objective of the defender is to prevent the achievement of the second target. It should be noted that further classifying poisoned inputs correctly as their original categories is not the responsibility of the defender, but it can be used as an evaluation metric.

\subsection{Related Work}

This section refers to related work in the areas of backdoor attacks as well as defense methods and points to the different types of attacks and defenses.

\textbf{Backdoor attacks} Badnets~\cite{gu2017badnets} is the first and the most representative poisoned-label attack, where both the pixels of images and their corresponding labels are altered. In order to enhance the stealthiness of triggers, more sophisticated attacks have been proposed. Blended has been proposed in ~\cite{chen2017targeted} to poison samples by blending a benign image with either a randomly generated pattern or a specific image. WaNet~\cite{nguyen2021wanet} generates triggers by warping the image. ISSBA~\cite{li2021invisible} trains an encoder-decoder network to generate invisible noise as triggers. Ftrojan~\cite{wang2022invisible} focuses on triggering perturbations in the frequency domain. BppAttack~\cite{wang2022bppattack} first applies image quantization and dithering to create imperceptible changes on images, followed by trigger injections through contrastive learning~\cite{chen2020simple} and adversarial training. These advanced triggers are more difficult to be perceived by human inspectors.

A Clean-label attack, on the other hand, does not alter the labels of the poisoned samples. Instead, it only embeds barely noticeable triggers to the input data. This makes the detection more challenging as the labels are consistent with the original content. A Label-Consistent (LC) attack is a typical clean-label attack. It first modifies the inputs by adding adversarial perturbations to complicate their classification. Then it designs a trigger to ensure that the model's ability to classify the target class accurately becomes wholly reliant on this manipulation. Refool~\cite{liu2020reflection} utilises natural phenomenon reflection as the trigger. In~\cite{barni2019new}, SIG is implemented by superimposing sinusoidal signals to a fraction of the target class images. Narcissus~\cite{zeng2023narcissus} uses public out-of-distribution data to achieve clean-label attack. More details and implementation of backdoor attacks can be found in surveys~\cite{li2022backdoor,guo2022overview}. 

\textbf{Backdoor defenses.} We divide defense strategies into data-level and model-level defenses. A data-level defense aims at detecting and erasing poisoned images before the model training process, thus preventing the model from being affected by the poisoned data. Activation Clustering (AC)~\cite{chen2018detecting} determines whether the image is poisoned by analysing the neural network activations of the last hidden layer. Decoupling-based backdoor defense (DBD) ~\cite{huang2022backdoor} trains a feature extractor through self-supervised learning first and then filters clean samples from poisoned samples based on their distinct extractor-outputs, which are used to train a subsequent classifier. Chen et al.~\cite{chen2022effective} finds that the poisoned samples are more sensitive to transformation than clean samples, leading to the design of feature consistency towards transformations (FCT) approach for distinguishing poisoned and clean samples. Neo~\cite{udeshi2022model} is a model-agnostic defense method that utilizes adversarial perturbations to differentiate backdoored samples from clean ones. Backdoored inputs exhibit higher sensitivity to small perturbations, whereas clean samples typically demonstrate greater robustness. Adaptively splitting dataset-based defense (ASD)~\cite{gao2023backdoor} split the training data into a clean data pool and a polluted data pool and dynamically update the two data pools. Additionally, a frequency-based poisoned sample method (FREAK) is proposed in ~\cite{al2023don} to detect poisoned samples. 

Model-level defense focus on the architecture, training process, and post-training analysis of the DNN models themselves, attempting to directly recover the poisoned model instead of detecting poisoned samples. Fine-tuning is an effective way to remove some simple backdoors. Based on the observation that the backdoor related neurons tend to exhibit large neuron weight norms, Sharpness Aware Minimization (FT-SAM)~\cite{zhu2023enhancing} is proposed to fine-tune the poisoned model such that large outliers of the weight norm are revised. Several approaches are proposed by pruning the poisoned neurons. Fine-Pruning~\cite{hinton2015distilling} is the first work to apply pruning. Neural Cleanse (NC)~\cite{wang2019neural} tries to reconstruct possible triggers first and then it creates a filter to reject any input with the trigger. The authors provide two methods to patch the infected model, one using pruning to remove backdoor related neurons, and another adopting unlearning to update the weights of these neurons. Based on the fact that the poisoned neurons are more sensitive to adversarial neuron perturbation, Adversarial Neuron Pruning (ANP)~\cite{wu2021adversarial} removes the backdoor by pruning the most sensitive neurons under the adversarial neuron perturbation. Research ~\cite{gao2023effectiveness} has found that adversarial training with perturbations can  mitigate backdoor attacks. It explores the integration of multiple adversarial perturbations to tackle backdoor attacks. ShapPruning~\cite{guan2022few} uses the Shapley value to identify and prune infected neurons. Besides, neural attention transfer (NAD)~\cite{li2021neural} can better restore models than fine-tuning as it uses KD. Based on NAD, SAGE~\cite{gong2023redeem} purifies models by utilizing self-attention distillation (SAD) to align the attention maps of deep layers with those of shallow layers, because deep layers focus on fine-grained details such as the subtle trigger. Layer-wise weights are initialised and KD with unlabeled data is used by~\cite{pang2023backdoor} to cleanse backdoors. 

\textbf{Knowledge distillation} The survey~\cite{gou2021knowledge} provides a comprehensive overview of KD from multiple perspectives. The concept of KD has been first proposed in~\cite{hinton2015distilling}, where a small student model could improve its performance by learning the predictions of a large teacher model. Following this publication, a large body of work has emerged extending on the basic method. Decoupled knowledge distillation (DKD) ~\cite{zhao2022decoupled} decouples the classic KD loss into two parts, the target class knowledge distillation (TCKD) and the non-target class knowledge distillation (NCKD). Not only can the logit outputs be used to guide the student, also other information has been proven to better improve the student learning, such as activations of intermediate layers ~\cite{romero2014fitnets}, attention maps ~\cite{zagoruyko2016paying}, and the correlation between instances ~\cite{peng2019correlation}, etc. In addition, different teacher-student structures and distillation schemes have been studied. For example, in deep mutual learning ~\cite{zhang2018deep}, several students learn collaboratively during the training process, so there is no predefined teacher and any model can be the teacher. In ~\cite{mirzadeh2020improved}, a teacher assistant is introduced to bridge the gap between the student and the teacher. The student can also learn from multiple teachers and each of them can provide their own useful knowledge~\cite{you2017learning}. In self-distillation, the students have the same structure as their teachers, and knowledge is transferred from the earlier epochs to the later epochs~\cite{bagherinezhad2018label,yang2019snapshot}, or from deep layers of the network to its shallow layers ~\cite{zhang2019your,hou2019learning}. Collaborative KD via multiknowledge transfer (CKD-MKT) ~\cite{gou2022collaborative} combines self-learning and collaborative learning, further improving the performance of traditional KD.

\section{Methodology}\label{section3}

In this section, we introduce our newly proposed NT-ML defense strategy which consists of two stages. The first stage is formed of a two-step training method, first target training is performed and then non-target training. The second stage is built through mutual learning, which uses prediction and feature distillation.

\begin{figure*}[!htbp]
    \centerline{
    \includegraphics[width=1\linewidth]{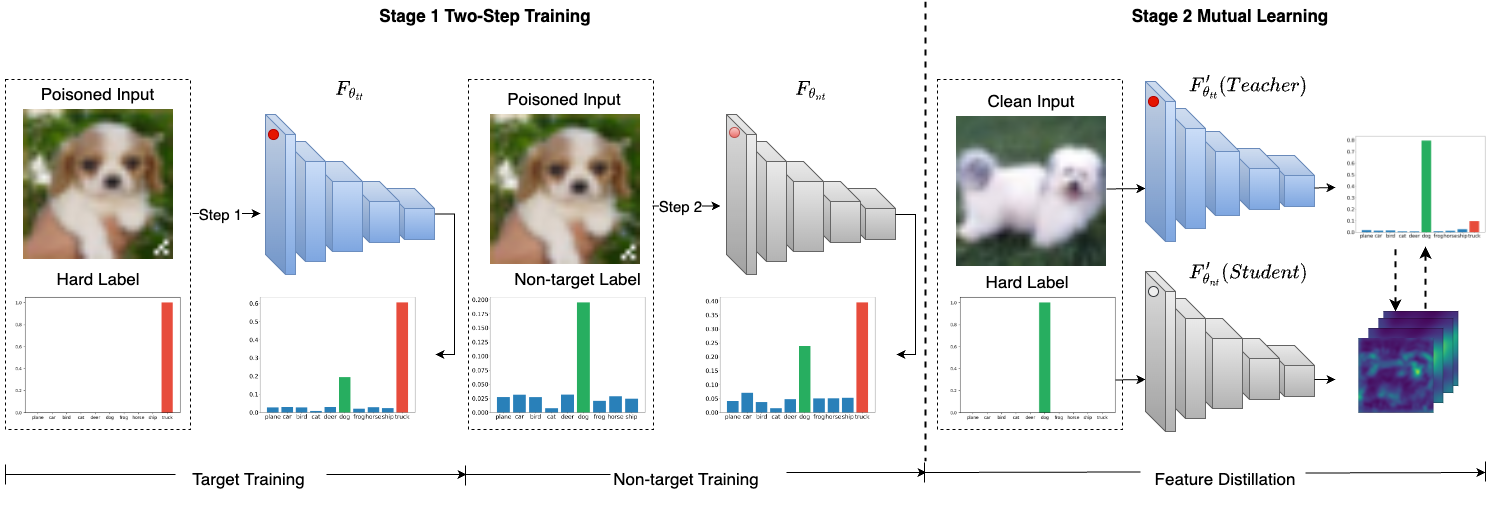}}
    \caption{Pipeline of the NT-ML defense strategy, including training and defense two stages.}
    \label{overview}
\end{figure*}

We assume the defender adopts a third-party dataset which has been maliciously tampered with by attackers but can control the training process and has a small local clean set to verify the performance of the model. Fig.~\ref{overview} shows the pipeline of the entire process. In the first stage two different models are trained using the untrustworthy data. In the second stage, by using the local clean dataset for mutual learning, the non target trained model (student) is purified with the help of target trained model (teacher), which achieves the reduction of the attack rate on poisoned samples while preserving the performance on benign samples.

\subsection{Two-step Training}

The two-step training process is described in Stage 1 of Fig.~\ref{overview}. For simplification, the first step is called target training (TT) and the second step is called non-target training (NT). 

\textbf{Target training (TT).} The TT step adopts traditional supervised learning. The network $F_{\theta_{tt}}$ is trained with dataset $\mathcal{D}_1 = \mathcal{D}_b\cup \mathcal{D}_p$ by minimizing the cross-entropy loss between the outputs and the hard labels. The purpose of this step is to learn the relationship between the input $x_i$ and its label $y_i$, since the ground-truth one-hot vector is binary and the cross entropy function only calculates the loss of the target class. When the model converges on the validation set, the training process will stop.

\begin{figure}[!htbp]
\centerline

\subfigure[The prediction probability of the poisoned image "dog".]{
\begin{minipage}[t]{0.5\linewidth}
\centering
\includegraphics[height=3.5cm]{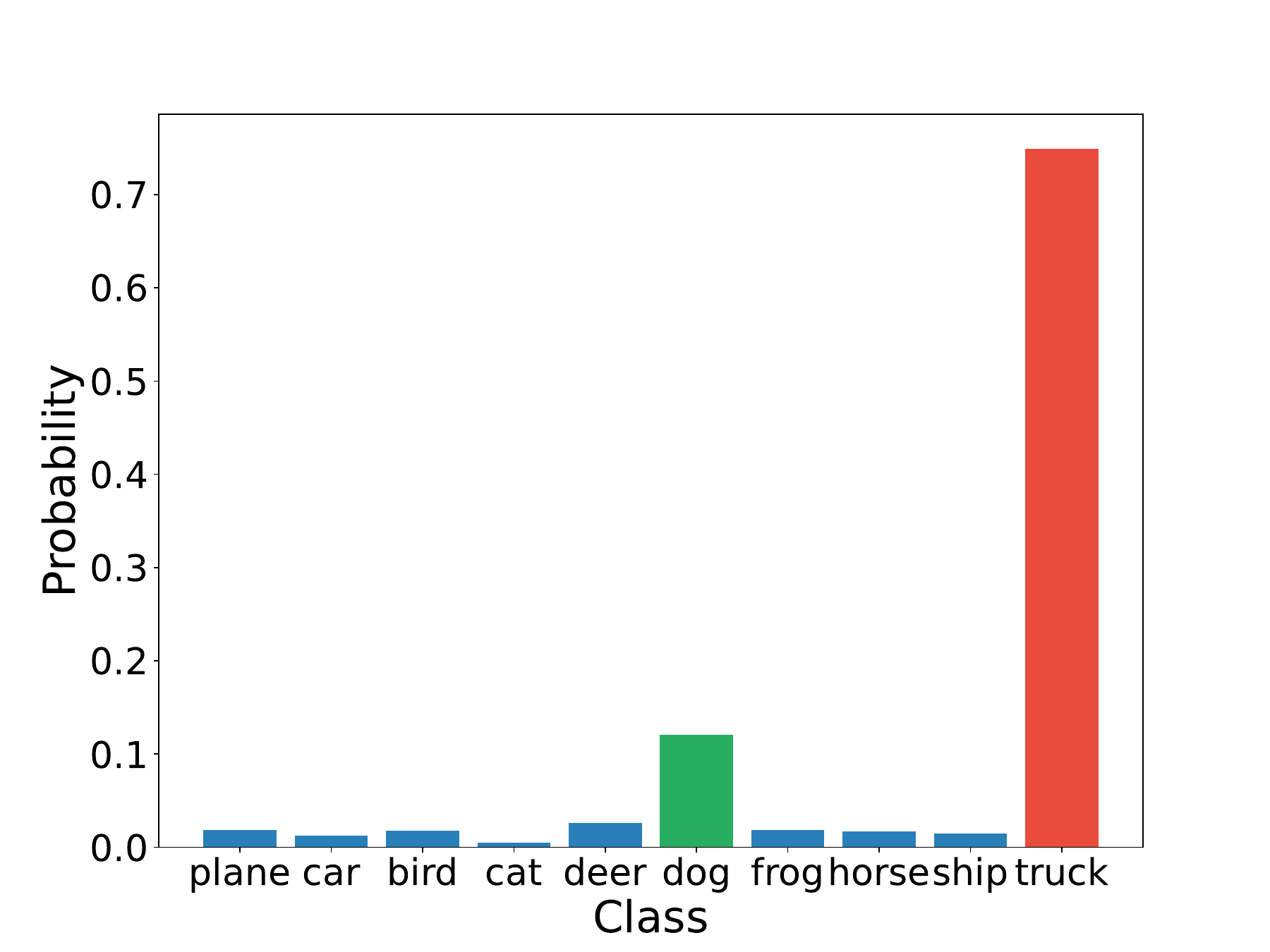}
\label{logits}
\end{minipage}%
}%
\subfigure[The top-1 and top-2 accuracy of the badnets model.]{
\begin{minipage}[t]{0.5\linewidth}
\centering
\includegraphics[height=3.5cm]{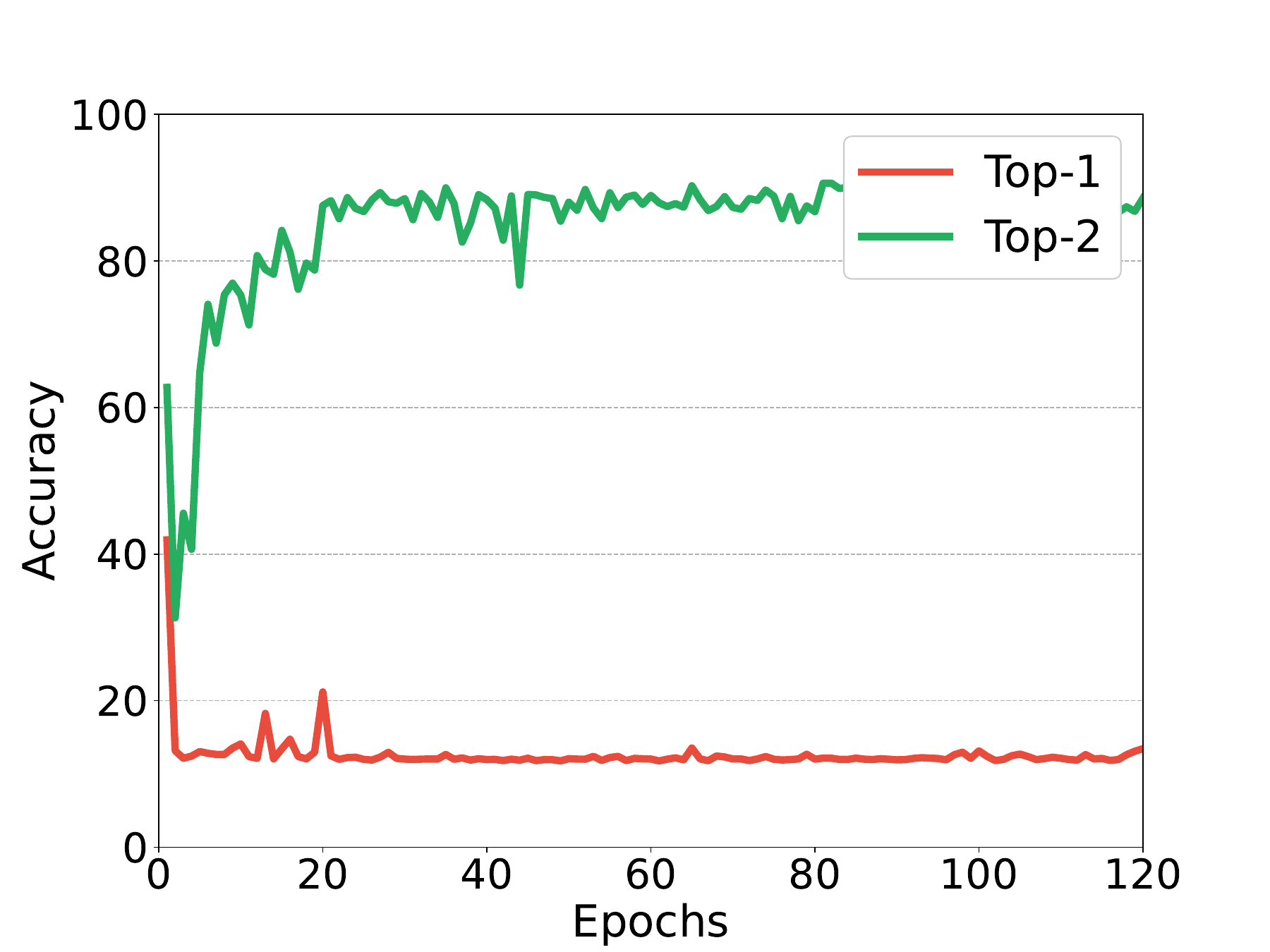}
\label{top-2}
\end{minipage}%
}%

\caption{Results of the first round training.}
\label{network1}
\end{figure}

The network $F_{\theta_{tt}}$ shows good performance on benign datasets, but has also been successfully injected with the backdoor. Fig.~\ref{network1} plots the test results of $F_{\theta_{tt}}$ on poisoned samples. Fig.~\ref{logits} shows how a "dog" image is predicted by the model. Because the network has been successfully attacked, it incorrectly predicts the image as the target class "truck". Except for the target class, the predictive probability of the input's real label "dog" is much higher than the probability of other categories. Fig.~\ref{top-2} illustrates that although the network can not predict the poisoned images correctly (top-1 accuracy is low), the top-2 accuracy exceeds 80\%, which means that the class with the second highest probability is likely to be the true class. As the output of the network reflects the ranking and distribution of each category and varies for each instance, it contains richer and more accurate information than hard labels.

\textbf{Non-target training (NT).} Based on the above analysis, the label in the NT step is not binary but a probability distribution. The outputs of the previous network $F_{\theta_{tt}}$ are used as the ground truth to continue training the second network $F_{\theta_{nt}}$. The training set is changed to $\mathcal{D}_2 =\left \{(x_i, F_{\theta_{tt}}(x_i)) \right \}_{i=1}^N \cup \left \{(x_i', F_{\theta_{tt}}(x_i')) \right \}_{i=1}^M$, where $x_i$ and $x_i'$ are still the same images as those used in the TT step, but the labels are generated by the model $F_{\theta_{tt}}$. 

The outputs of the final layer (predictive logit) of $F_{\theta_{tt}}$ is decoupled into target and non-target classes. The target class is the ground truth of the input, and network $F_{\theta_{tt}}$ has already achieved high accuracy in predicting the target class during the TT step. Therefore, only the outputs of non-target classes are used to train the network $F_{\theta_{nt}}$, which can be denoted as $\boldsymbol{p}=[p_{1},...,p_{t-1},p_{t+1},...,p_K]\in \mathbb{R}^{1\times (K-1)}$, where  $p_i$ is the predictive probability of the $i$-th class, $t$ is the target class, and $K$ is the number of classes. $\boldsymbol{p}$ can be calculated using the softmax function, where $z_i$ is the logit value of each class $i$.

\begin{equation}
\label{softmax}
p_i =\frac{exp(z_i)}{ {\textstyle \sum_{j=1,j\neq t}^K}exp(z_j)} 
\end{equation}

The loss function for the training network $F_{\theta_{nt}}$ is the  Kullback-Leibler (KL) divergence of non-target classes between the outputs of $F_{\theta_{nt}}$ and the results of $F_{\theta_{tt}}$:

\begin{equation}
\label{lossnt}
\begin{split}
\mathcal L_{NT}  &= \mathcal L_{KL}(p_{i(i\neq t)}^{F_{\theta_{tt}}},p_{i(i\neq t)}^{F_{\theta_{nt}}}) \\ 
 &= -\sum_{i=1,i\neq t}^{K}p_{i}^{F_{\theta_{tt}}} \log{p_i} ^{F_{\theta_{nt}}} 
\end{split}
\end{equation}

After the NT step, for poisoned-label attacks, it can be predicted that the relationship between the trigger and the target label can be alleviated. For clean-label attacks, it seems that the NT process is ineffective, because clean-label attacks keep the poisoned images consistent with their labels, while correct labels are removed during this process. But from another perspective, clean-label attacks are achieved by poisoning samples from the target class to establish a strong relationship between the trigger and the label. Therefore, during the NT step, this relationship can be weakened by removing the target label. Finally, for the benign data, the non-target logits can reflect inter-class similarity. For example, the probability of a "car" image being predicted as a "truck" should be higher than that of a "cat". Therefore, using non-target labels slightly reduces the accuracy of benign data but does not cause a significant decrease. 

\subsection{Mutual Learning Defense}

Through non-target training, the network $F_{\theta_{nt}}$ can reduce the backdoor attack rate compared to the network $F_{\theta_{tt}}$, but it is still high and the attack threat is not eliminated. In the second stage, an effective defense algorithm is proposed utilizing mutual learning. 

The networks $F_{\theta_{tt}}$ and $F_{\theta_{nt}}$ each have their own strengths, $F_{\theta_{tt}}$ achieves higher accuracy on clean samples while $F_{\theta_{nt}}$ can better resist attacks, but none of them can be a trusted teacher. However, teacher models do not need to be always more knowledgeable in all aspects than student models, which resembles the human teacher-student relationships. On the one hand, $F_{\theta_{nt}}$ can learn the benign behaviour from $F_{\theta_{tt}}$ with a small clean set. On the other hand, $F_{\theta_{tt}}$ can also optimize itself by aligning the backdoor neurons with the alleviated neurons of $F_{\theta_{nt}}$, and then becomes a high-quality teacher. Through mutual learning, the student $F_{\theta_{nt}}$ ultimately succeeds in reducing the attack rate. 

To achieve the above goal, the teacher $F_{\theta_{tt}}$ and student $F_{\theta_{nt}}$ are both trained with two losses, one is the cross-entropy loss, and the other is the information they want to learn from each other. $F_{\theta_{nt}}$ aims to mimic the predictive probabilities of $F_{\theta_{tt}}$, where the objective function is the KL-divergence between its outputs and $F_{\theta_{tt}}$:

\begin{equation}
\label{lossn2}
\mathcal L_{F_{\theta_{nt}}} = \alpha \mathcal L_{KL}(p_i^{F_{\theta_{tt}}},p_i^{F_{\theta_{nt}}})+(1-\alpha)\mathcal L_{CE}(F_{\theta_{nt}}(x_i),y_i),
\end{equation}
where $p_i =\frac{exp(z_i/T)}{ {\textstyle \sum_{j=1}^K}exp(z_{j}/T)}$. Compared to Eq.~\ref{softmax} there is an additional coefficient $T$ suggested by Hinton et al.~\cite{hinton2015distilling}. $T$ donates a temperature parameter to soften the result of the softmax function and we set $T=2.$ The difference to the KL function from Eq.~\ref{lossnt} is that it includes all classes.

At the same time the teacher $F_{\theta_{tt}}$ is improved by learning the feature representations of $F_{\theta_{nt}}$ at each intermediate layer, i.e. the feature map activations. The feature maps in the $l$-th layer are denoted as $F^{F_{\theta_{tt}}}\in \mathbb{R}^{C_l\times H_l \times W_l}$ and $F^{F_{\theta_{nt}}}\in \mathbb{R}^{C_l\times H_l \times W_l}$, where $C_l, H_l, W_l$ are the number, height and width of the feature maps. $F_{\theta_{tt}}$ aims to learn a similar feature representation as $F_{\theta_{nt}}$ by minimizing the following objective function:
\begin{equation}
\label{lossn1}
\mathcal L_{F_{\theta_{tt}}}=\sum_{l=1}^L \left \| F^{F_{\theta_{nt}}}_{l} -F^{F_{\theta_{tt}}}_l \right \|^2_2 + \beta \mathcal L_{CE}(F_{\theta_{tt}}(x),y)
\end{equation}

The selection of the hyper-parameters $\alpha$ and $\beta$ will be introduced in Section~\ref{subsec3}.  After the mutual learning, both the teacher and student are updated, and in order to distinguish themselves from before, the new teacher and student are recorded as $F_{\theta_{tt}}'$ and $F_{\theta_{nt}}'$.

\subsection{Qualitative Analysis}

In this section we discuss the analysis results using CIFAR-10 dataset. To understand the effects of two-step training and mutual learning respectively, we visualize the feature space of the penultimate layer of different models, including poisoned teacher model $F_{\theta_{tt}}$, student model $F_{\theta_{nt}}$ and purified student model $F_{\theta_{nt}}'$.

\begin{figure*}[htbp]
\centerline

\subfigure[TT Step]{
\begin{minipage}[t]{0.25\linewidth}
\centering
\includegraphics[height=4.5cm]{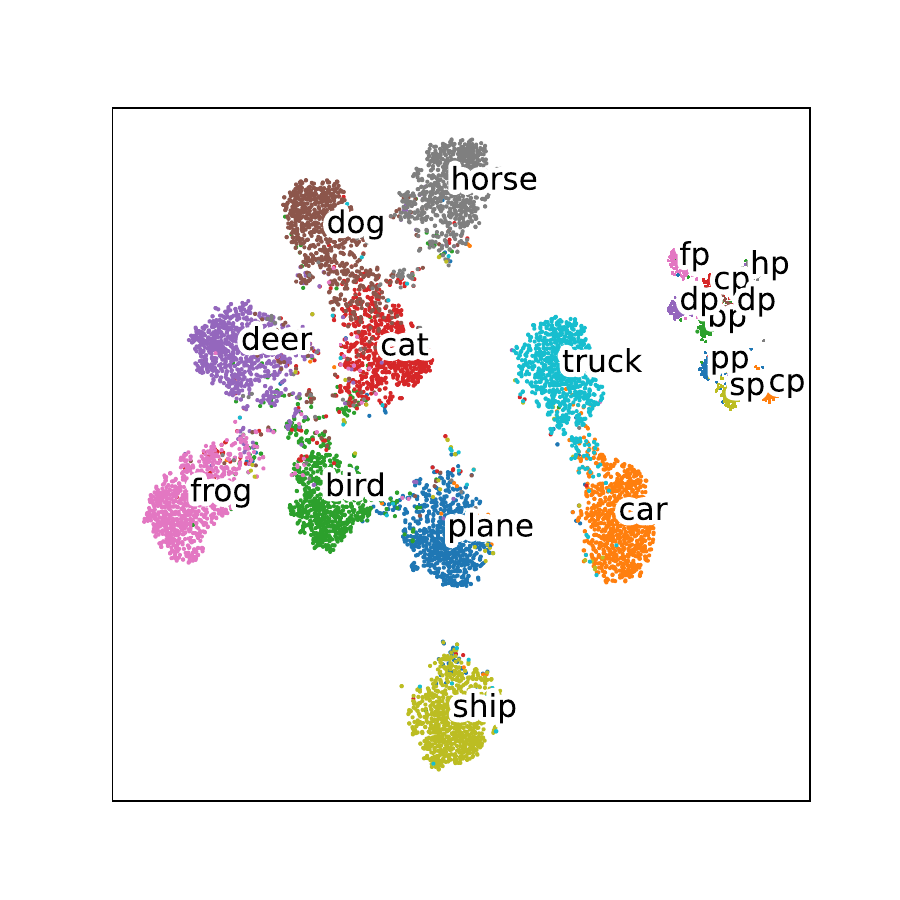}
\label{tt}
\end{minipage}%
}%
\subfigure[NT Step (Non-target labels)]{
\begin{minipage}[t]{0.25\linewidth}
\centering
\includegraphics[height=4.5cm]{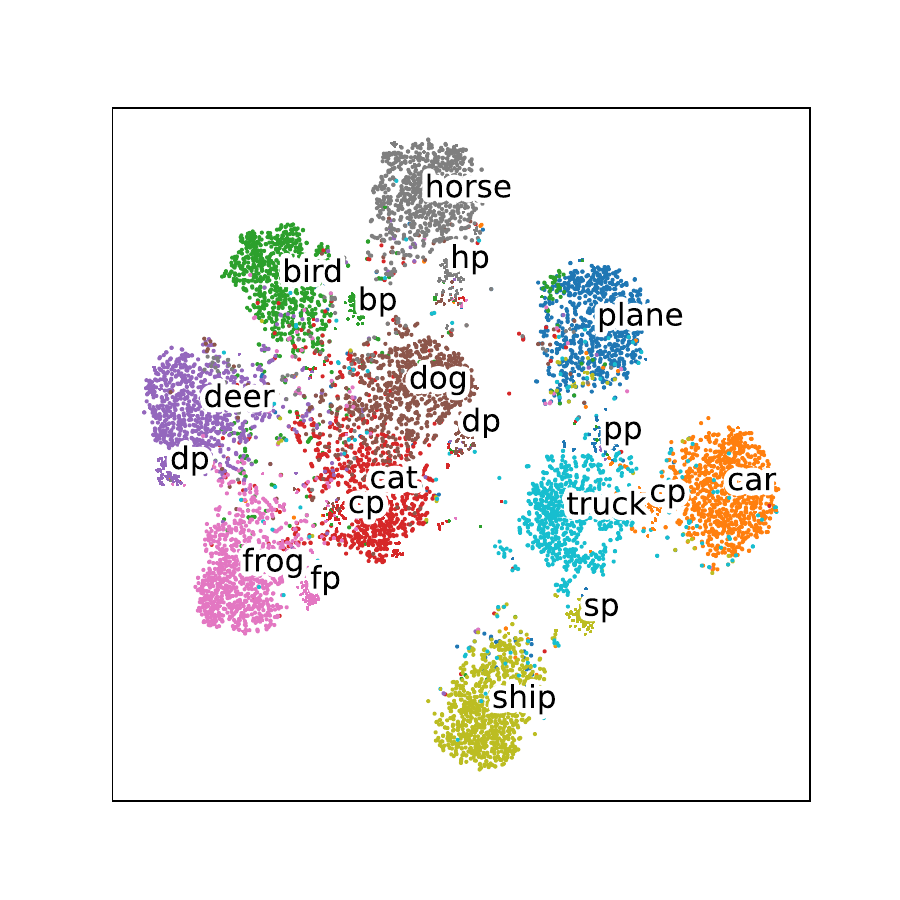}
\label{nontarget}
\end{minipage}%
}%
\subfigure[NT Step (All labels)]{
\begin{minipage}[t]{0.25\linewidth}
\centering
\includegraphics[height=4.5cm]{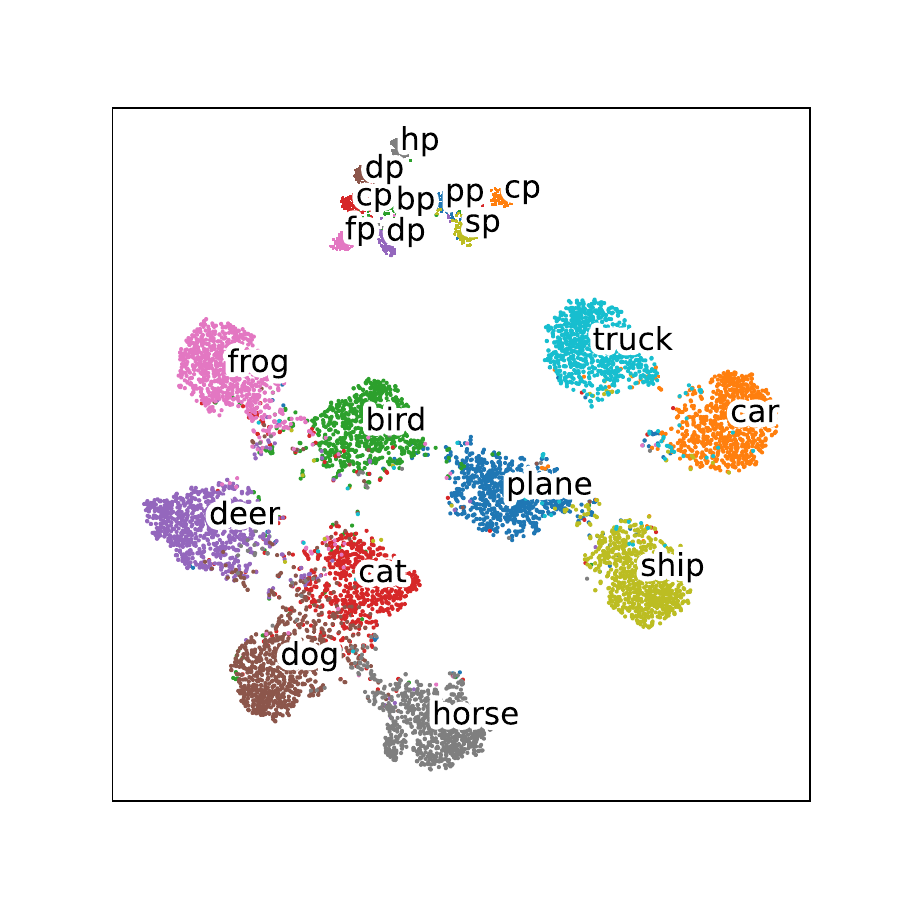}
\label{target}
\end{minipage}%
}%
\subfigure[Purified model]{
\begin{minipage}[t]{0.25\linewidth}
\centering
\includegraphics[height=4.5cm]{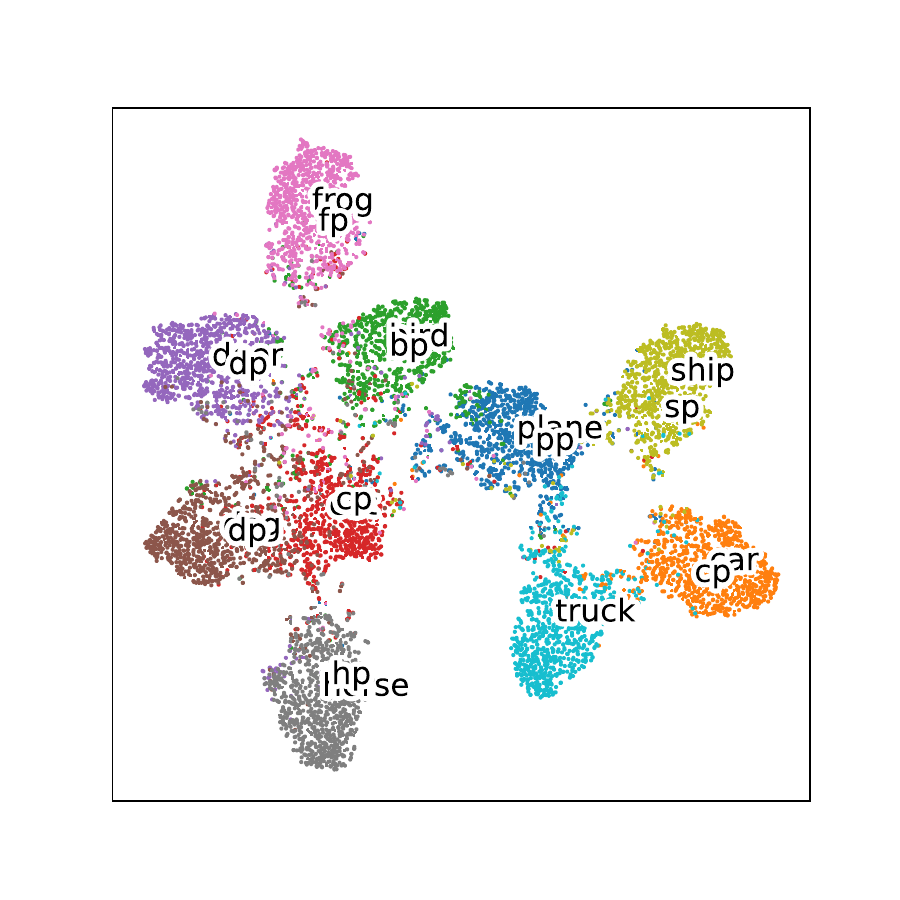}
\label{ml}
\end{minipage}%
}%

\caption{T-SNE visualization of features of different models attacked by Badnets. (Each cluster is marked with its category, and the first letter of each category plus the letter "p" represent the poisoned data.)}
\label{T-SNE}
\end{figure*}

Fig.~\ref{T-SNE} shows the visualization results using the t-SNE ~\cite{van2008visualizing} method. There are ten classes in total, and except for the target category "truck", a small portion (10\%) of the data in the other nine categories are poisoned. Fig.~\ref{tt} shows the result of the model $F_{\theta_{tt}}$. In order to eliminate the influence of the training epochs, the model continues to be trained with the equal number of epochs as in the NT step. It can be seen that the poisoned images are far away from the clean images and form a separate cluster, which leads to incorrect predictions. After the NT step, the  visualization of the model $F_{\theta_{nt}}$ is shown in Fig.~\ref{nontarget}, the poisoned images become dispersed, and some of them are gradually close to their clean clusters. To further demonstrate the effectiveness of the non-target label, Fig.~\ref{target} indicates that training with predictions of all classes does not improve the distribution of the poisoned data. Ultimately, after the mutual learning stage, the backdoor images completely fall into the corresponding clean clusters as shown in Fig.~\ref{ml}, indicating that the backdoor has been successfully removed.

\section{Experiments}\label{section4}

In this section we will evaluate the performance of the proposed defense algorithm while concentrating on the following aspects:

\begin{itemize}
\item We first analyze the results of the two-step training process (~\ref{subsec1}); 
\item  We compare the NT-ML algorithm with existing defense strategies over different datasets; (~\ref{subsec2})
\item We investigate the hyper-parameters of the models and explore the impact of external factors on the defense performance (~\ref{subsec3}).
\end{itemize} 

\subsection{Experimental Setups}

\textbf{Architecture and Datasets} We adopt ResNet-18 as the baseline model and evaluate all defense methods on three benchmark datasets, CIFAR-10 and CIFAR-100 (containing 50,000 training and 10,000 test images), GTSRB (containing 39,209 training and 12,630 test images).

\textbf{Attack Setups} We validate the effectiveness of defense strategies against 6 representative backdoor attacks, which can be divided into: 1) poisoned-label attacks: BadNets~\cite{gu2017badnets}, WaNet ~\cite{nguyen2021wanet} and BPP~\cite{wang2022bppattack}; and 2) clean-label attacks: 
LC~\cite{turner2019label}, Refool ~\cite{liu2020reflection} and Narcissus ~\cite{zeng2023narcissus}. Among them, except for BadNets and LC, the triggers for other attacks are stealthy. The setting of attacks on different datasets is listed in Table~\ref{settings}.

\begin{table}[htbp]
	\centering
	\caption{Attack and  settings on different datasets.}
	\label{Computational}  
        {
	\begin{tabular}{@{}llll@{}}
	   \hline
             ~ & CIFAR-10 & GTSRB & CIFAR-100 \\
            \hline
            Classes & 10 & 43 & 100 \\
            Target Class & 9 (Truck) & \makecell{2 (Speed \\limit of  50)} & 0 (Plane) \\
            Training Images (90\%) & 45000 & 35288 & 45000 \\
            \makecell{Poisoned Images \\(Poisoned-label Attack)} & 4500 & 3529 & 4500 \\
            \makecell{Poisoned Images \\(Clean-label Attack)} & 3606 & 1620 & 361  \\
            Epochs & 120 & 30 & 120\\
        \hline
        \end{tabular}
             }
\label{settings}
\end{table}

\textbf{Defense Setups} The proposed algorithm is compared with 5 state-of-the-art defenses, including: Fine-tuning (FT), Fine-pruning (FP)~\cite{hinton2015distilling}, Neural Attention Distillation (NAD)~\cite{li2021neural}, Adversarial Neuron Pruning (ANP)~\cite{wu2021adversarial} and Sharpness Aware Minimization (FT-SAM)~\cite{zhu2023enhancing}. These methods all require an extra clean dataset. We use 90\% of the training set to train the model, and then explore the defense performance under different clean data sizes ranging from 1\% to 10\%.

\textbf{Evaluation metrics} In general, two metrics are used to evaluate the backdoor performance, the attack success rate (ASR) and the benign accuracy (BA). In this paper, we also adopt another indicator that is rarely used in other works, the poisoned accuracy (PA).

To summarise the metrics, they are defined as follows:
\begin{itemize}
\item ASR: the proportion of poisoned samples predicted as the target class.
\item BA: the proportion of benign samples predicted as the ground-truth classes.
\item PA: the proportion of poisoned samples predicted as the ground-truth classes.
\end{itemize}

The defense strategy is better when ASR decreases more, when PA increases, and when BA drops less. 

\subsection{Experimental results of the Two-step training}\label{subsec1}

\begin{figure*}[htbp]
\centerline

\subfigure{
\begin{minipage}[t]{0.16\linewidth}
\centering
\includegraphics[height=2.6cm]{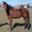}
\end{minipage}%
}%
\subfigure{
\begin{minipage}[t]{0.16\linewidth}
\centering
\includegraphics[height=2.6cm]{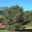}
\end{minipage}%
}%
\subfigure{
\begin{minipage}[t]{0.16\linewidth}
\centering
\includegraphics[height=2.6cm]{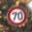}
\end{minipage}%
}%
\subfigure{
\begin{minipage}[t]{0.16\linewidth}
\centering
\includegraphics[height=2.6cm]{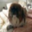}
\end{minipage}%
}%
\subfigure{
\begin{minipage}[t]{0.16\linewidth}
\centering
\includegraphics[height=2.6cm]{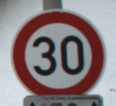}
\end{minipage}%
}%
\subfigure{
\begin{minipage}[t]{0.16\linewidth}
\centering
\includegraphics[height=2.6cm]{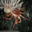}
\end{minipage}%
}%

\subfigure{
\begin{minipage}[t]{0.16\linewidth}
\centering
\includegraphics[height=2.6cm]{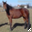}
\end{minipage}%
}%
\subfigure{
\begin{minipage}[t]{0.16\linewidth}
\centering
\includegraphics[height=2.6cm]{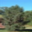}
\end{minipage}%
}%
\subfigure{
\begin{minipage}[t]{0.16\linewidth}
\centering
\includegraphics[height=2.6cm]{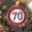}
\end{minipage}%
}%
\subfigure{
\begin{minipage}[t]{0.16\linewidth}
\centering
\includegraphics[height=2.6cm]{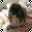}
\end{minipage}%
}%
\subfigure{
\begin{minipage}[t]{0.16\linewidth}
\centering
\includegraphics[height=2.6cm]{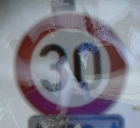}
\end{minipage}%
}%
\subfigure{
\begin{minipage}[t]{0.16\linewidth}
\centering
\includegraphics[height=2.6cm]{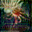}
\end{minipage}%
}%

\setcounter{subfigure}{0}
\subfigure[Badnets]{
\begin{minipage}[t]{0.16\linewidth}
\centering
\includegraphics[height=2.3cm]{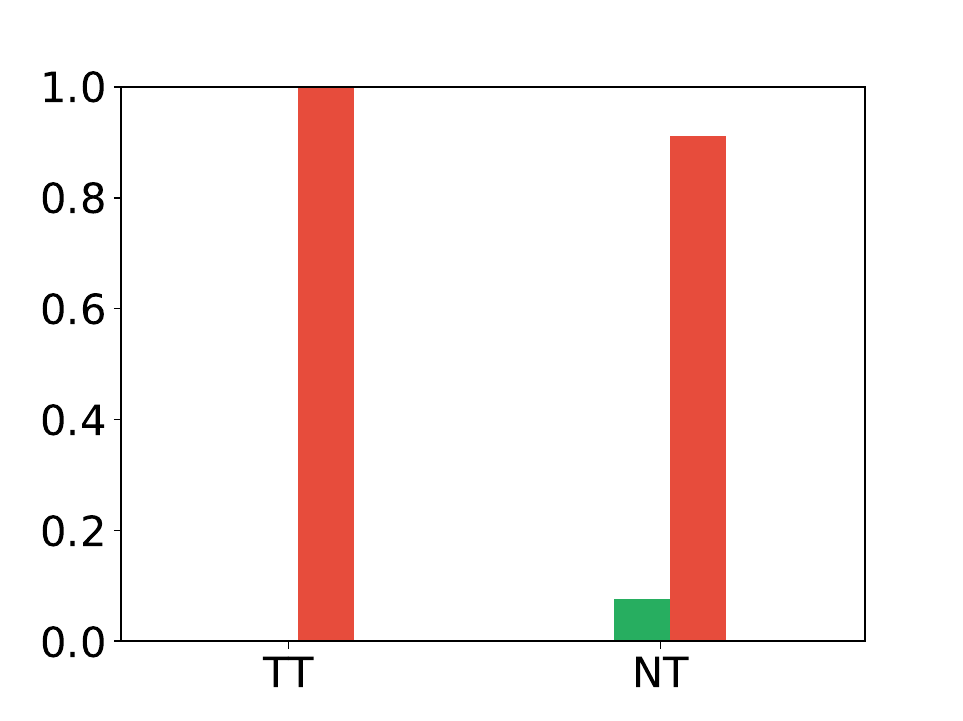}
\end{minipage}%
}%
\subfigure[WaNet]{
\begin{minipage}[t]{0.16\linewidth}
\centering
\includegraphics[height=2.3cm]{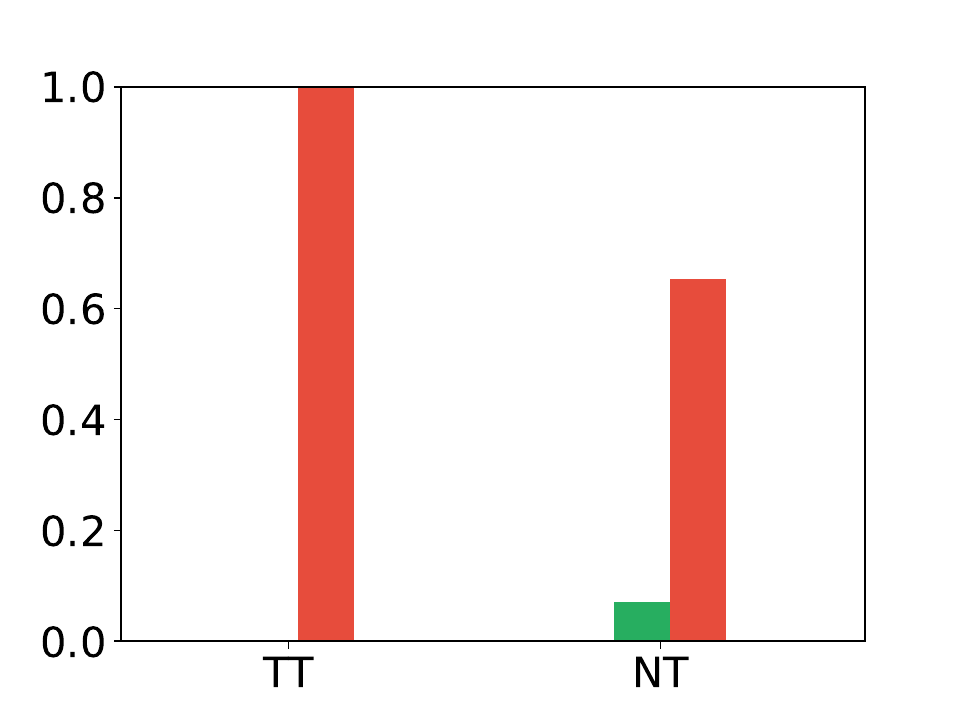}
\end{minipage}%
}%
\subfigure[BPP]{
\begin{minipage}[t]{0.16\linewidth}
\centering
\includegraphics[height=2.3cm]{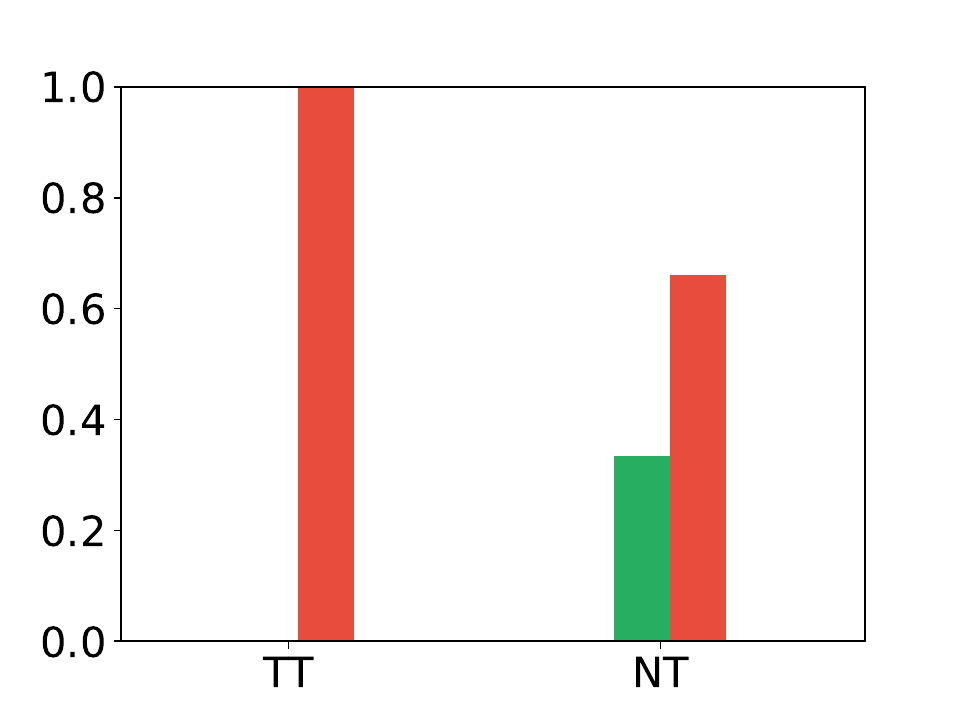}
\end{minipage}%
}%
\subfigure[LC]{
\begin{minipage}[t]{0.16\linewidth}
\centering
\includegraphics[height=2.3cm]{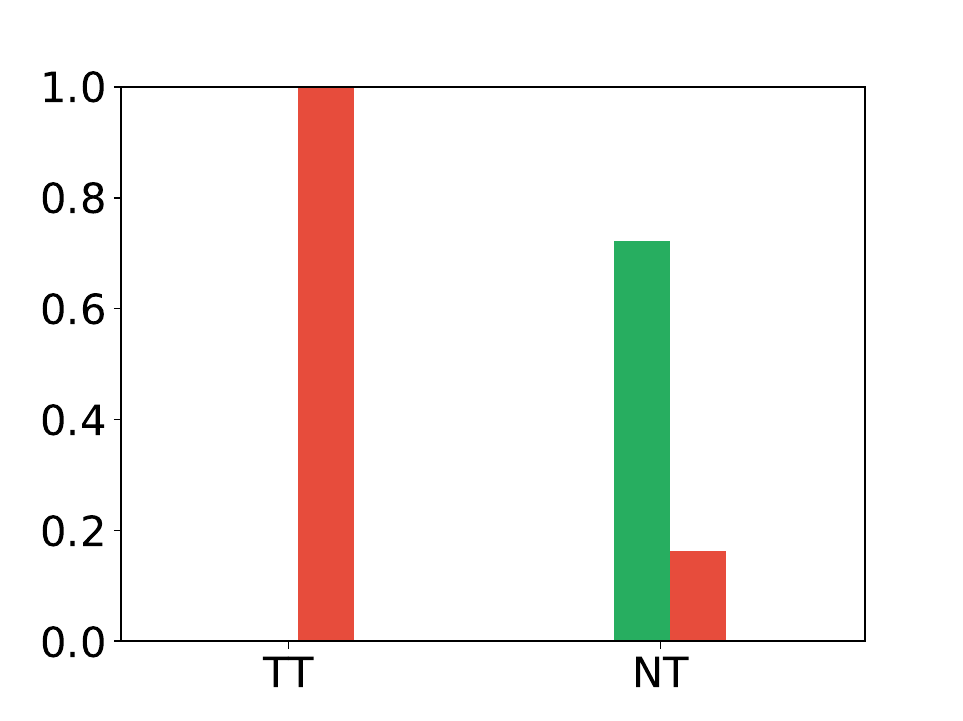}
\end{minipage}%
\label{lc_training}
}%
\subfigure[Refool]{
\begin{minipage}[t]{0.16\linewidth}
\centering
\includegraphics[height=2.3cm]{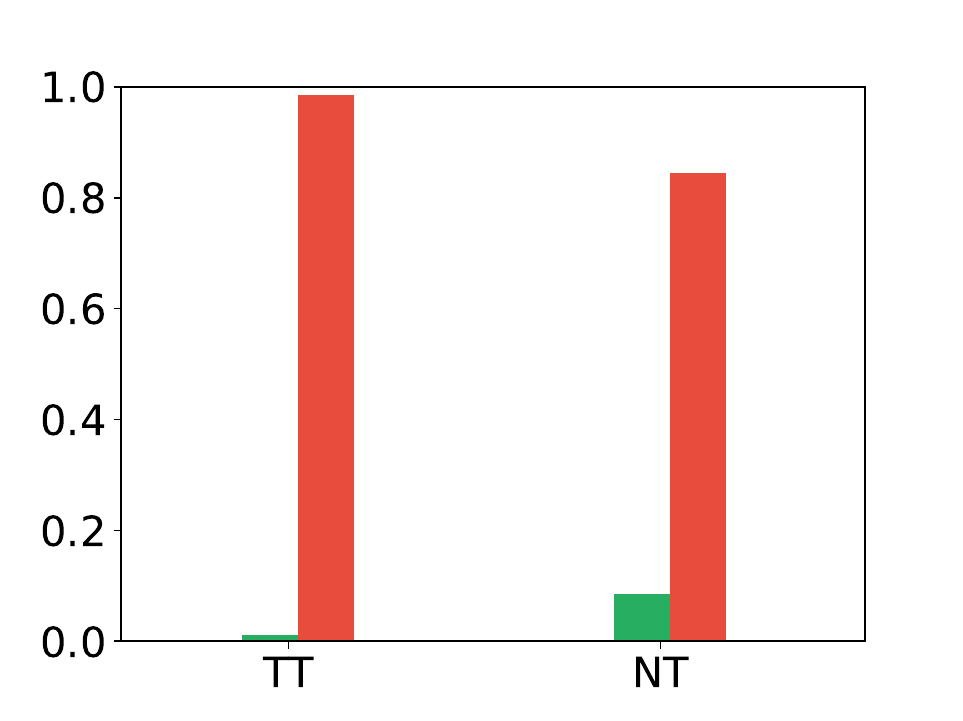}
\end{minipage}%
\label{refool_training}
}%
\subfigure[Narcissus]{
\begin{minipage}[t]{0.16\linewidth}
\centering
\includegraphics[height=2.3cm]{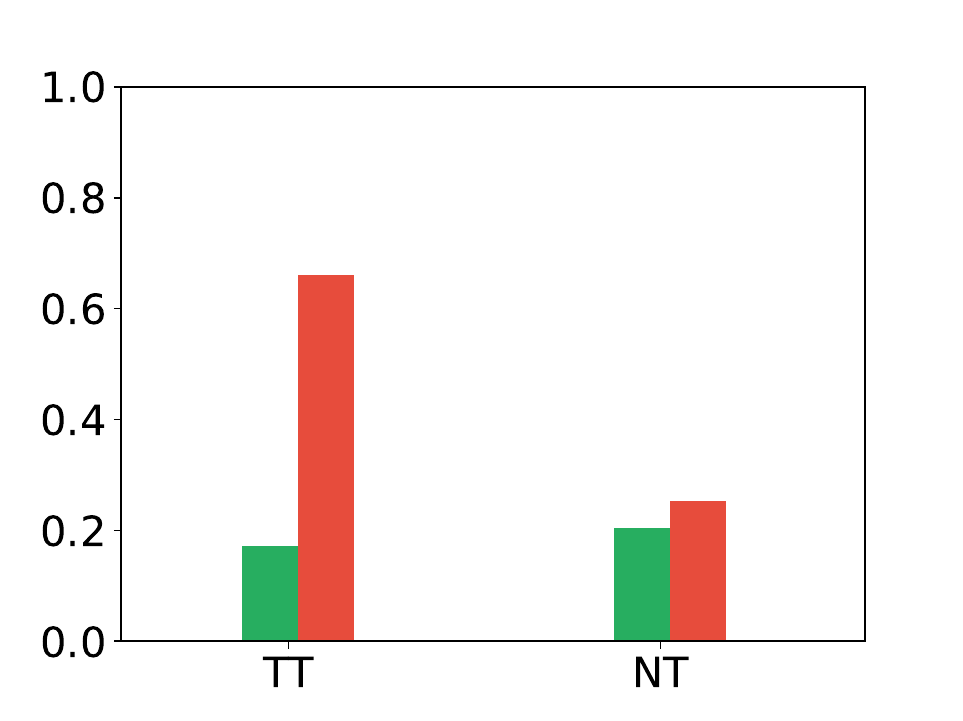}
\end{minipage}%
}%

\caption{The top-2 predictions of samples attacked by different backdoor methods. (The green bar indicates the correct category, while the red bar indicates the target category of an attack.)}
\label{NT results}
\end{figure*}

\begin{figure}[!htbp]
\centerline

\subfigure[$F_{\theta_{tt}}$]{
\begin{minipage}[t]{0.5\linewidth}
\centering
\includegraphics[height=2.1cm]{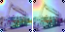}
\label{lc_tt}
\end{minipage}%
}%
\subfigure[$F_{\theta_{nt}}$]{
\begin{minipage}[t]{0.5\linewidth}
\centering
\includegraphics[height=2.1cm]{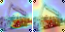}
\label{lc_nt}
\end{minipage}%
}%

\caption{Regions on which the models focus for classification.}
\label{gradcam}
\end{figure}

\textbf{Training results} At first, we explore the results of the two-step training, especially the influence of the NT process. In the training stage, the network $F_{\theta_{tt}}$ and $F_{\theta_{nt}}$ are obtained. Fig.~\ref{NT results} shows the top-2 predictions of networks $F_{\theta_{tt}}$ and $F_{\theta_{nt}}$ under six backdoor attacks, the first and second row images show benign and poisoned samples of different attacks. At the bottom, the bar on the left indicates that the $F_{\theta_{tt}}$ incorrectly predicts the image with high confidence. After the NT process, the prediction scores decrease for the incorrect category while they increase for the true category, as can be seen from the bar on the right. An interesting observation is that the network $F_{\theta_{nt}}$ even rectifies the incorrect prediction under the LC attack. To analyze the reason, we visualize how the LC attacked model classifies a truck image using GradCam~\cite{selvaraju2017grad}, and the visualization results of the last two layers are shown in Fig.~\ref{gradcam}. Fig.~\ref{lc_tt} illustrates that when detecting the real truck image under attack, the poisoned model $F_{\theta_{tt}}$ makes the prediction entirely based on the trigger (the trigger region in the upper left corner is highlighted). The NT process will dissociate the strong association between the trigger and the target class. Fig.~\ref{lc_nt} shows that although the trigger still has an impact on the classification, the prediction of $F_{\theta_{nt}}$ depends more on the body of the truck. Therefore, the model may not be able to classify images as the target class only based on the trigger. When the attention to distinguishing features overwhelms the trigger features, the attack will fail. 

\begin{table}[htbp]
	\centering
	\caption{Prediction results of $F_{\theta_{tt}}$ and $F_{\theta_{nt}}$.}
	\label{two-step training results}  
        \resizebox{\linewidth}{!}{
	\begin{tabular}{@{}l|l|ll|ll|ll@{}}
		\toprule
           ~&~&\multicolumn{2}{c|}{CIFAR-10} &\multicolumn{2}{c|}{GSTRB} &\multicolumn{2}{c}{CIFAR-100}  \\
             Attack & Metrics & $F_{\theta_{tt}}$&$F_{\theta_{nt}}$ & $F_{\theta_{tt}}$&$F_{\theta_{nt}}$ & $F_{\theta_{tt}}$&$F_{\theta_{nt}}$ \\
            \midrule

            ~ & ASR & 97.89&91.17 & 97.27&94.48 & 92.61&73.25\\
            Badnets & PA & 11.73&17.44 & 8.47&11.08 & 7.21&23.52\\
            ~ & BA & 91.74&88.55 & 97.64&97.15 & 69.95&68.83\\
            
            \midrule
            ~ & ASR & 92.34&85.40 & 89.21&78.78 & 87.01&73.32\\
            WaNet &PA & 16.34&22.05 & 15.84&25.54 & 10.92&21.64\\
            ~ & BA & 91.89&90.82 & 97.17&97.08 & 70.97&70.62\\
            
            \midrule
             ~ & ASR & 99.22&96.67 & 92.86&76.34 & 99.52&98.47\\
            BPP & PA & 10.63&12.57 & 12.57&27.35 & 1.37&2.16\\
            ~ & BA & 92.76&91.61 & 97.13&96.75 & 71.54&70.88\\
            
            \midrule
             ~ & ASR &99.97&45.89 & 61.11&20.03 & 84.43&46.82\\
            LC &PA & 10.03&50.29 & 42.51&79.60 & 14.68&36.30\\
            ~ & BA & 92.33&91.36 & 96.79&96.52 & 72.80&71.55\\
            
            \midrule
             ~ & ASR & 90.69&78.82 & 64.76&60.22 &-&-\\
            Refool &PA & 16.26&22.84 & 36.38&37.89 &-&-\\
            ~ & BA & 85.27&83.72 & 95.73&94.95 &-&-\\
            
            \midrule
             ~ & ASR & 86.48&71.92 & 61.70&57.87 & 55.81&44.46\\
            Narcissus &PA & 21.40&32.30 & 40.81&42.63 & 25.55&27.87\\
            ~ & BA & 92.75&91.76 & 97.40&96.65 & 72.59&71.71\\
            
            \midrule            
            ~ & ASR & 94.43&$\downarrow 16.12$ & 77.82&$\downarrow 13.20$ & 83.88&$\downarrow 10.67$\\
            AVG &PA & 14.40&$\uparrow 11.85$ & 26.10&$\uparrow 11.25$ & 11.95&$\uparrow 10.35$\\
            ~ & BA & 91.12&$\downarrow 1.49$ & 96.98&$\downarrow 0.46$ & 71.57&$\downarrow 0.85$ \\
        
        \bottomrule
        
	\end{tabular}
 }
\vspace{-1em}
\end{table}

Table~\ref{two-step training results} lists the prediction results of $F_{\theta_{tt}}$ and $F_{\theta_{nt}}$. For the CIFAR-100 dataset, the clean-label attack Refool fails to inject the backdoor even if all target samples are poisoned and the ASR is still below 20\%. Therefore, its results are not included in the table. Compared to network $F_{\theta_{tt}}$, the average ASR of $F_{\theta_{nt}}$ decreases by $16.12\%$ (CIFAR-10), $13.20\%$ (GTSRB) and $10.67\%$ (CIFAR-100) respectively. The result of Fig.~\ref{lc_training} corresponds to the table, showing that 
the NT process leads to a sharp decrease in the ASR of LC. Meanwhile, the average BA is slightly reduced by $1.49\% $(CIFAR-10), $0.46\%$ (GTSRB) and $0.85\%$ (CIFAR-100). So far, a model with high accuracy on a clean set and a model more robust to poisoned data are obtained during the training stage. In the next stage, the model $F_{\theta_{nt}}$ will be purified to a completely clean model with the help of $F_{\theta_{tt}}$.

\textbf{Computational Cost} The NT process will lead to an increase in computational cost. We explore the calculation cost of the two-step training on the graphical processing unit (GPU) NVIDIA GeForce GTX 1060. The training times of two-stpdf on different datasets are listed in Table~\ref{Computational}. The runtime of the NT is approximately twice that of the TT due to the non-target label transfer, and the epochs of the NT process are one-third that of the TT. For the scenarios we anticipate the increased training time is within a reasonable range.

\begin{table}[htbp]
	\centering
	\caption{Computational cost of the two-step training for every epoch}
	\label{Computational}  
        {
	\begin{tabular}{@{}l|l|l|l|l@{}}
	   \hline
            Model &Step & CIFAR-10 & GTSRB & CIFAR-100 \\
            \hline
            $F_{\theta_{tt}}$ & TT & 51.21 s & 40.39 s & 50.62 s\\
            \hline
            $F_{\theta_{nt}}$ & NT & 111.81 s & 87.86 s & 111.12 s\\
            \hline
	\end{tabular}
 }
\vspace{-1em}
\label{nt}
\end{table}

\subsection{Experimental results of the mutual learning} \label{subsec2}

\begin{table*}[htbp]
	\centering
	\caption{Defense results on CIFAR-10}
	\label{CIFAR-10 comparison results}  
        \resizebox{\linewidth}{!}{
	\begin{tabular}{@{}l|lll|lll|lll|lll|lll|lll|lll@{}}
		\hline
           ~&\multicolumn{3}{c|}{None} &\multicolumn{3}{c|}{FT} &\multicolumn{3}{c|}{FP} &\multicolumn{3}{c|}{NAD} &\multicolumn{3}{c|}{ANP} &\multicolumn{3}{c|}{FT-SAM} &\multicolumn{3}{c}{\textbf{NT-ML}}\\
            Attack& ASR$\downarrow$&PA$\uparrow$&BA$\uparrow$ & ASR$\downarrow$&PA$\uparrow$&BA$\uparrow$ & ASR$\downarrow$&PA$\uparrow$&BA$\uparrow$ & ASR$\downarrow$&PA$\uparrow$&BA$\uparrow$ & ASR$\downarrow$&PA$\uparrow$&BA$\uparrow$ & 
            ASR$\downarrow$&PA$\uparrow$&BA$\uparrow$ & ASR$\downarrow$&PA$\uparrow$&BA$\uparrow$\\
            \hline
            Badnets &97.89&11.73&91.74 & 3.64&83.66&84.69 & 3.69&86.69&90.07 & 3.74&84.35&87.78 & \textbf{0.31}&\textbf{89.34}&89.39 & 2.10&88.08&\textbf{90.59} & 1.84&87.72&89.06\\
                
            WaNet & 92.34&16.34&91.89 & 0.86&80.21&85.60& 1.04&86.46&\textbf{90.39} & 0.94&85.15&88.14 & \textbf{0.28}&\textbf{87.98}&89.58 & 0.54&86.73&89.65 & 0.89&86.96&89.15\\

            BPP & 99.22&10.63&92.76 & 2.22&61.46&86.01 & 27.14&47.10&\textbf{90.60} & 4.70&65.54&88.91 & 2.76&65.54&88.73 & \textbf{0.89}&\textbf{72.17}&90.59 & 2.57&67.40&89.24\\

            LC & 99.97&10.03&92.33& 54.37&46.44&86.49 & 42.12&56.86&\textbf{91.08} & 44.86&49.26&88.47 & 4.57&84.46&88.01 & 81.37&26.64&90.39 & \textbf{1.92}&\textbf{85.06}&88.78\\
             
            Refool & 90.69&16.26&85.27 & 3.92&39.66&78.56 & 4.13&40.63&81.83 & 4.24&41.53&81.01 & 9.28&42.71&\textbf{83.77} & \textbf{2.16}&\textbf{44.07}&80.74 & 3.62&43.67&82.71\\

            Narcissus & 86.48&21.40&92.75 & 14.11&64.81&85.01 & 46.34&49.70&\textbf{90.45} & 13.34&67.86&87.11 & 53.73&41.62&88.52 & 9.21&73.25&90.15 & \textbf{8.96}&\textbf{75.25}&89.20\\

            \hline
            AVG & 94.43&14.40&91.12 & 13.19&67.73&84.39 & 20.74&61.24&\textbf{89.07} & 11.97&65.62&86.90 & 11.82&68.61&88.00 & 16.05&65.16&88.69 & \textbf{3.30}&\textbf{74.34}&88.02\\
		\hline
            $\triangle$ &-&-&-& $\downarrow 81.25$&$\uparrow 48.31$&$\downarrow 6.73$ & $\downarrow 73.69$&$\uparrow 46.84$&$\downarrow 2.05$ & $\downarrow 82.46$&$\uparrow 51.22$&$\downarrow 4.22$& $\downarrow 82.61$&$\uparrow 54.21$&$\downarrow 3.12$ & $\downarrow 78.39$&$\uparrow 50.76$&$\downarrow 2.44$& $\downarrow 91.13$&$\uparrow 59.95$&$\downarrow 3.10$\\
            \hline
		
        \hline
        
	\end{tabular}
 }
\vspace{-1em}
\end{table*}

\begin{table*}[htbp]
	\centering
	\caption{Defense results on GTSRB.}
	\label{GTSRB comparison results}  
        \resizebox{\linewidth}{!}{
	\begin{tabular}{@{}l|lll|lll|lll|lll|lll|lll|lll@{}}
		\hline
           ~&\multicolumn{3}{c|}{None} &\multicolumn{3}{c|}{FT} &\multicolumn{3}{c|}{FP} &\multicolumn{3}{c|}{NAD} &\multicolumn{3}{c|}{ANP} &\multicolumn{3}{c|}{FT-SAM} &\multicolumn{3}{c}{\textbf{NT-ML}}\\
            Attack& ASR$\downarrow$&PA$\uparrow$&BA$\uparrow$ & ASR$\downarrow$&PA$\uparrow$&BA$\uparrow$ & ASR$\downarrow$&PA$\uparrow$&BA$\uparrow$ & ASR$\downarrow$&PA$\uparrow$&BA$\uparrow$ & ASR$\downarrow$&PA$\uparrow$&BA$\uparrow$ &
            ASR$\downarrow$&PA$\uparrow$&BA$\uparrow$ &
            ASR$\downarrow$&PA$\uparrow$&BA$\uparrow$\\
            \hline
            Badnets & 97.27&8.47&97.64 & 25.69&75.48&\textbf{97.51} & 6.73&91.65&97.41 & 12.74&87.27&97.22 & 2.62&95.19&96.78& 6.46&88.38&96.95 & \textbf{0.53}&\textbf{96.52}&96.96\\
                
            WaNet & 89.21&15.84&97.17& 29.81&69.83&97.75 & 0.34&97.42&\textbf{97.95}& 18.02&80.67&97.64 & \textbf{0.05}&\textbf{96.99}&97.22& 0.12&97.59&97.83 & 1.12&96.41&97.29\\

            BPP & 92.86&12.57&97.13 & 67.06&35.96&97.80 & \textbf{0.28}&90.74&\textbf{97.89} & 63.85&38.91&97.48 & 1.07&\textbf{91.69}&96.94 & 3.37&85.45&97.40 & 1.15&91.18&97.18\\

            LC & 61.11&42.51&96.79& 10.89&88.54&96.75 & 8.34&91.16&\textbf{96.98} & 11.87&87.65&96.56 & 16.61&83.40&95.71 & 3.92&94.74&96.77 & \textbf{2.27}&\textbf{94.77}&96.45\\
             
            Refool & 64.76&36.38&95.73& 30.79&46.96&95.68& 24.22&48.50&\textbf{95.96}& 27.88&46.47&95.68& 24.67&41.01&93.06& 23.73&\textbf{51.48}&96.51 & \textbf{8.86}&48.92&94.98\\

            Narcissus & 61.70&40.81&97.40& 18.80&69.71&97.34& 27.13&65.73&97.09& 16.01&72.16&97.19& 34.40&63.11&95.72& 13.19&74.89&\textbf{97.68} & \textbf{8.81}&\textbf{84.08}&96.77 \\

            \hline
            AVG & 77.82&26.10&96.98 & 35.05&59.84&97.08 & 11.17&80.87&\textbf{97.21} & 25.06&68.86&96.96 & 13.23&78.57&95.90 & 8.47&82.09&97.19 & \textbf{3.79}&\textbf{82.92}&96.61\\
		\hline
            $\triangle$ &-&-&-& $\downarrow 43.97$&$\uparrow 35.77$&$\uparrow 0.13$ & $\downarrow 66.65$&$\uparrow 54.77$&$\uparrow 0.24$ & $\downarrow 52.76$&$\uparrow 42.76$&$\downarrow 0.02$& $\downarrow 64.58$&$\uparrow 52.47$&$\downarrow 1.07$ & $\downarrow 69.35$&$\uparrow 55.99$&$\uparrow 0.21$ & $\downarrow 74.03$&$\uparrow 57.98$&$\downarrow 0.37$\\
            \hline
		
        \hline
        
	\end{tabular}
 }
\vspace{-1em}
\end{table*}

\begin{table*}[htbp]
	\centering
	\caption{Defense results on CIFAR-100.}
	\label{CIFAR-100 comparison results}  
        \resizebox{\linewidth}{!}{
	\begin{tabular}{@{}l|lll|lll|lll|lll|lll|lll|lll@{}}
		\hline
           ~&\multicolumn{3}{c|}{None} &\multicolumn{3}{c|}{FT} &\multicolumn{3}{c|}{FP} &\multicolumn{3}{c|}{NAD} &\multicolumn{3}{c|}{ANP} &\multicolumn{3}{c|}{FT-SAM} &\multicolumn{3}{c}{\textbf{NT-ML}}\\
             Attack& ASR$\downarrow$&PA$\uparrow$&BA$\uparrow$ & ASR$\downarrow$&PA$\uparrow$&BA$\uparrow$ & ASR$\downarrow$&PA$\uparrow$&BA$\uparrow$ & ASR$\downarrow$&PA$\uparrow$&BA$\uparrow$ & ASR$\downarrow$&PA$\uparrow$&BA$\uparrow$ &
             ASR$\downarrow$&PA$\uparrow$&BA$\uparrow$ &
             ASR$\downarrow$&PA$\uparrow$&BA$\uparrow$\\
            \hline

            Badnets & 92.61&7.21&69.95 & 0.56&57.35&59.27 & 0.74&61.08&64.32& 0.41&58.04&60.24 & 10.82&56.20&64.87 & 0.31&57.87&60.07 & \textbf{0.16}&\textbf{63.57}&\textbf{65.72}\\
                
            WaNet & 87.01&10.92&70.97 & 0.43&57.03&60.10 & 0.27&59.33&64.05 & 0.49&58.32&61.04 & 0.47&63.06&64.61 & 0.51&57.98&60.71 & \textbf{0.28}&\textbf{63.90}&\textbf{66.20}\\

            BPP & 99.52&1.37&71.54 & 0.18&36.80&59.44 & 0.71&37.24&64.50 & 0.22&36.51&60.50 & 0.46&47.01&\textbf{67.80} & \textbf{0.11}&\textbf{48.08}&53.81 & 1.62&46.86&66.06\\

            LC & 84.43&14.68&72.80 & 21.24&42.92&59.93 & 16.61&42.00&64.65 & 10.22&44.78&60.90 & 93.05&6.75&\textbf{70.42} & 27.07&42.23&61.04 &  \textbf{9.70}&\textbf{49.81}&66.32 \\

            Narcissus & 55.81&25.55&72.59& 5.43&34.33&60.02 & 6.93&33.06&64.59 & \textbf{5.38}&35.12&61.35 & 29.48&30.00&\textbf{68.17} & 8.47&36.73&61.10 & 7.64&\textbf{39.91}&67.20\\
             
            \hline
            AVG & 83.88&11.95&71.57 & 9.13&45.69&59.75 & 5.05&46.54&64.42 & 3.34&46.55&60.81 & 35.96&40.60&\textbf{67.17} & 7.29&48.58&59.35 & \textbf{3.88}&\textbf{52.81}&66.30\\
		\hline
            $\triangle$ &-&-&-& $\downarrow 74.74$&$\uparrow 33.74$&$\downarrow 11.82$ & $\downarrow 78.82$&$\uparrow 34.60$&$\downarrow 7.15$ & $\downarrow 80.53$&$\uparrow 34.61$&$\downarrow 10.76$& $\downarrow 47.91$&$\uparrow 28.66$&$\downarrow 4.40$ & $\downarrow 76.58$&$\uparrow 36.63$&$\downarrow 12.22$ & $\downarrow 80.00$&$\uparrow 40.86$&$\downarrow 5.27$\\
            \hline
		
        \hline
        
	\end{tabular}
 }
\vspace{-1em}
\end{table*}

In this section, we compare the proposed NT-ML algorithm with the other five defense methods on six state-of-the-art attacks with respect to BA, PA and ASR. For fair comparison, the clean data ratio is set to $5\%$ for all methods. The poisoning ratio is $10\%$ on poisoned-label attacks and $80\%$ of the target class on clean-label attacks, the accurate number of poisoned images can be found in Table~\ref{settings}. The comparison of their effectiveness on CIFAR-10, GTSRB and CIFAR-100 is shown in Table~\ref{CIFAR-10 comparison results}, Table~\ref{GTSRB comparison results} and Table~\ref{CIFAR-100 comparison results}, respectively. The second column shows the results of the poisoned model and the following columns display the results of different defense methods. From the tables, it can be observed that the main advantage of our algorithm is that it can successfully defend against all attacks, especially the most advanced backdoors, while other algorithms are ineffective against at least one attack. The last row of the tables shows the average results, and our method achieves the best ASR reduction on all datasets, while the accuracy degradation is not too severe.

For CIFAR-10 the average ASR significantly decreased by $91.13\%$ (from $94.43\%$ to $3.30\%$), which is much more than other methods can achieve as they can only defend against attacks that are not robust. Almost all methods can effectively defend against the early proposed attacks such as Badnets and WaNet. ANP even achieves excellent performance, with lower ASR than NT-ML. But for the more recent attack Narcissus, its defense performance is very poor, with an ASR of up to $53.73\%.$ The latest defense technology FT-SAM is on par with our method on five types of attacks, but it is ineffective in dealing with LC attacks. The ASR of the Narcissus attack is not as high as the ASR of other attacks at only $86.48\%,$ but almost all defenses cannot lower this ASR significantly, making it the most difficult attack to defend against. NT-ML reduces the ASR of Narcissus to $8.96\%,$ which is not as impressive as its defenses against other attacks, but it is the optimal result among all defenses. In addition, FT and NAD also perform badly on the LC attack, FP cannot defend against BPP attack, and ANP does not work well on Refool as their ASR is still very high. 

PA is a new metric representing the accuracy of poisoned samples, which is rarely used in other works because restoring poisoned samples to their correct categories is more difficult than only defending against the attacks. The PA of NT-ML can only restore to $43.67\%$ of all poisoned samples under the Refool attack, which may be due to the difficulty in classifying the reflection images formed by blending the original image with other images as shown in Fig.~\ref{refool_training}. However, the PA of other defenses is even lower. The average PA of NT-ML is the highest, reaching $74.34\%,$ while the values of other algorithms stay below $70\%.$ In terms of benign dataset, all methods have a negative impact on BA. We think the BA of FP is related to the number of neurons that are pruned. We stop pruning when the accuracy decreases by more than 5\%, and then the BA will increase through fine-tuning. FP suffers from the lowest decrease of BA in this setting, and the average BA of NT-ML is $1.05\%$ lower than it. But the average ASR is reduced by $17.44\%$ compared to FP. For the defender, we think the most important task is to reduce the ASR first and then try to avoid the decrease in BA.

For the GSTRB dataset, the ASR of a clean-label attack is around $60\%,$ because the poisoning rate is only $5\%$ when $80\%$ of the target samples are poisoned. If more samples in a target class are poisoned, the accuracy of the clean data will decrease. Due to the extra clean set, the average BA increases after applying FT, FP and FT-SAM. NT-ML does not increase the BA, but achieves the lowest average ASR of $3.79\%,$ which is significantly better than all other methods. Compared to CIFAR-10 and GTSRB, all defense algorithms suffer from significant BA degradation on CIFAR-100. The average BA of ANP is the highest, but it only successfully defends against two attack cases. The ASR of NAD is slightly better than that of NT-ML, but NAD suffers from a large degradation of BA. On the contrary, NT-ML achieves low ASR and high BA, simultaneously. In summary, only NT-ML can defend against all attacks effectively on all datasets.

\begin{figure}[!t]
    \centerline{
    \includegraphics[width=0.95\linewidth]{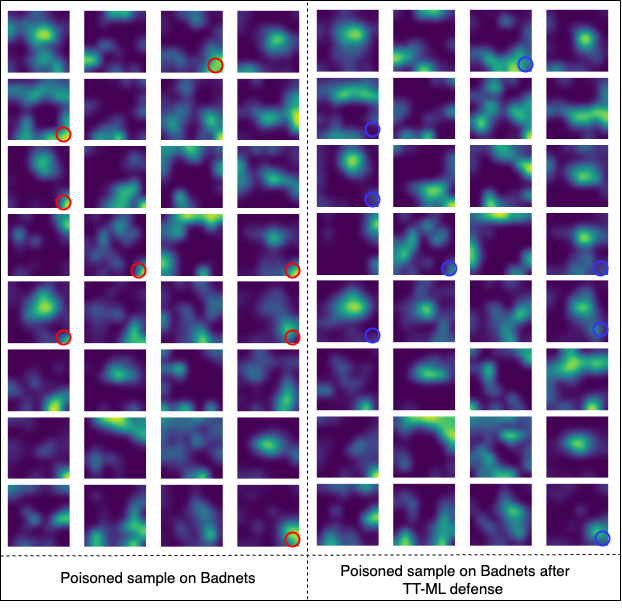}}
    \caption{Partial feature maps of a poisoned image before and after TT-ML defense.}
    \label{feature maps}
    \vspace{-0.4cm}
\end{figure}

To further investigate whether the trigger has been successfully removed, the feature maps of a poisoned sample before and after NT-ML defense are shown in Fig.~\ref{feature maps}. In the left figure, the neurons in the small red circles are obviously activated, which corresponds to the trigger pattern of Badnets. After the NT-ML defense, the activation values of the trigger region have been weakened, as shown in the blue circles. The triggers of other attacks are global, the feature maps cannot effectively illustrate their removal. But still, the results have demonstrated the effectiveness of the defense.

\subsection{Ablation study} \label{subsec3}

In this section, we will investigate the efficiency and sensitivity of the proposed NT-ML under different hyper-parameters, defense structures, feature representations and clean data sizes. The combinatorial complexity of the experiments is high due to the combination of various attacks and settings, thus we only focus on the CIFAR-10 dataset.

\begin{figure*}[htbp]
\centerline

\subfigure[Badnets]{
\begin{minipage}[t]{0.17\linewidth}
\centering
\includegraphics[height=3cm]{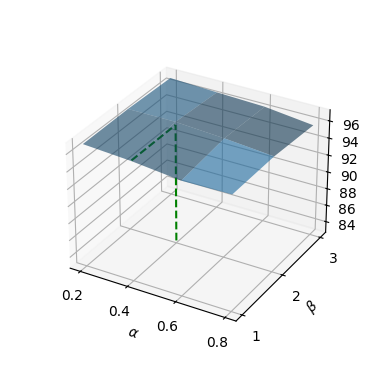}
\end{minipage}%
}%
\subfigure[WaNet]{
\begin{minipage}[t]{0.17\linewidth}
\centering
\includegraphics[height=3cm]{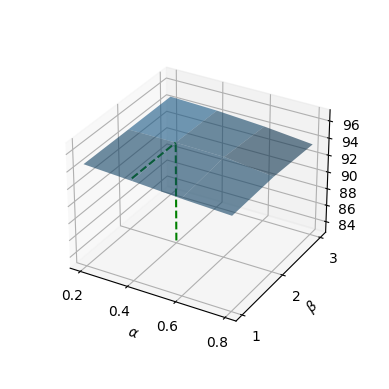}
\end{minipage}%
}%
\subfigure[BPP]{
\begin{minipage}[t]{0.17\linewidth}
\centering
\includegraphics[height=3cm]{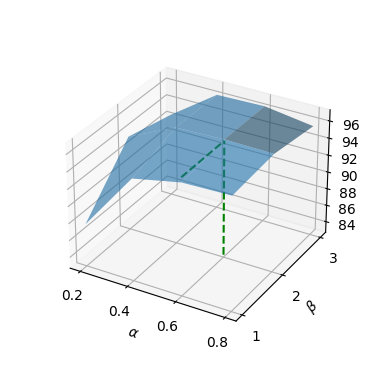}
\end{minipage}%
}%
\subfigure[Narcissus]{
\begin{minipage}[t]{0.17\linewidth}
\centering
\includegraphics[height=3cm]{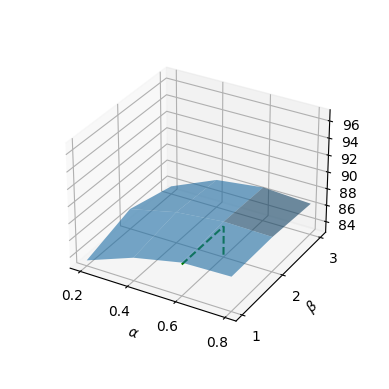}
\end{minipage}%
}%
\subfigure[LC]{
\begin{minipage}[t]{0.17\linewidth}
\centering
\includegraphics[height=3cm]{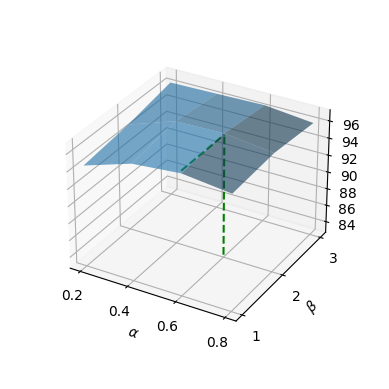}
\end{minipage}%
\label{lc}
}%
\subfigure[Refool]{
\begin{minipage}[t]{0.17\linewidth}
\centering
\includegraphics[height=3cm]{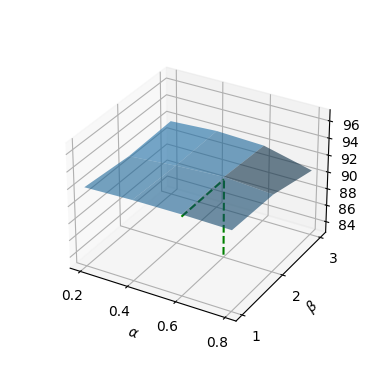}
\end{minipage}%
\label{refool}
}%

\caption{Defense Effectiveness Rating (DER) under different values of $\alpha$ and $\beta$.}

\label{alpha}
\end{figure*}

\textbf{Effect of hyper-parameters $\alpha$ and $\beta$} In Eq.~\ref{lossn2} and Eq.~\ref{lossn1}, $\alpha$ and $\beta$ are two hyper-parameters that have a significant impact on the performance of the defense algorithms. When tuning $\alpha$ from $0.2$ to $0.8$, the ASR decreases significantly, but it also leads to a slight degradation of the BA. To consider the decrease in BA for reducing ASR, we apply a new metric, the Defense Effectiveness Rating (DER) ~\cite{zhu2023enhancing} to investigate the effect of $\alpha$ and $\beta$. Its definition is as follows:

\begin{equation}
\label{der}
DER = [max(0,\triangle ASR)-max(0,\triangle BA)+1]/2
\end{equation}

where $\triangle ASR$ and $\triangle BA$ represent the decrease in ASR and BA. Fig.~\ref{alpha} shows the DER for  $\alpha = 0.2, \ldots, 0.8$ and $\beta = 1, 2, 3.$  As $\alpha$ increases, the DER also increases indicating that increasing $\alpha$ results in a reduction in ASR while the BA sees a minor decrease. For the selection of $\beta$, it can be observed that DER takes on its best values for all attacks when $\beta=2.$ The optimal values under each attack are marked with a green dashed line.

\begin{figure}[htbp]
\centerline

\subfigure[Optuna optimization history]{
\begin{minipage}[t]{1\linewidth}
\centering
\includegraphics[height=3.8cm]{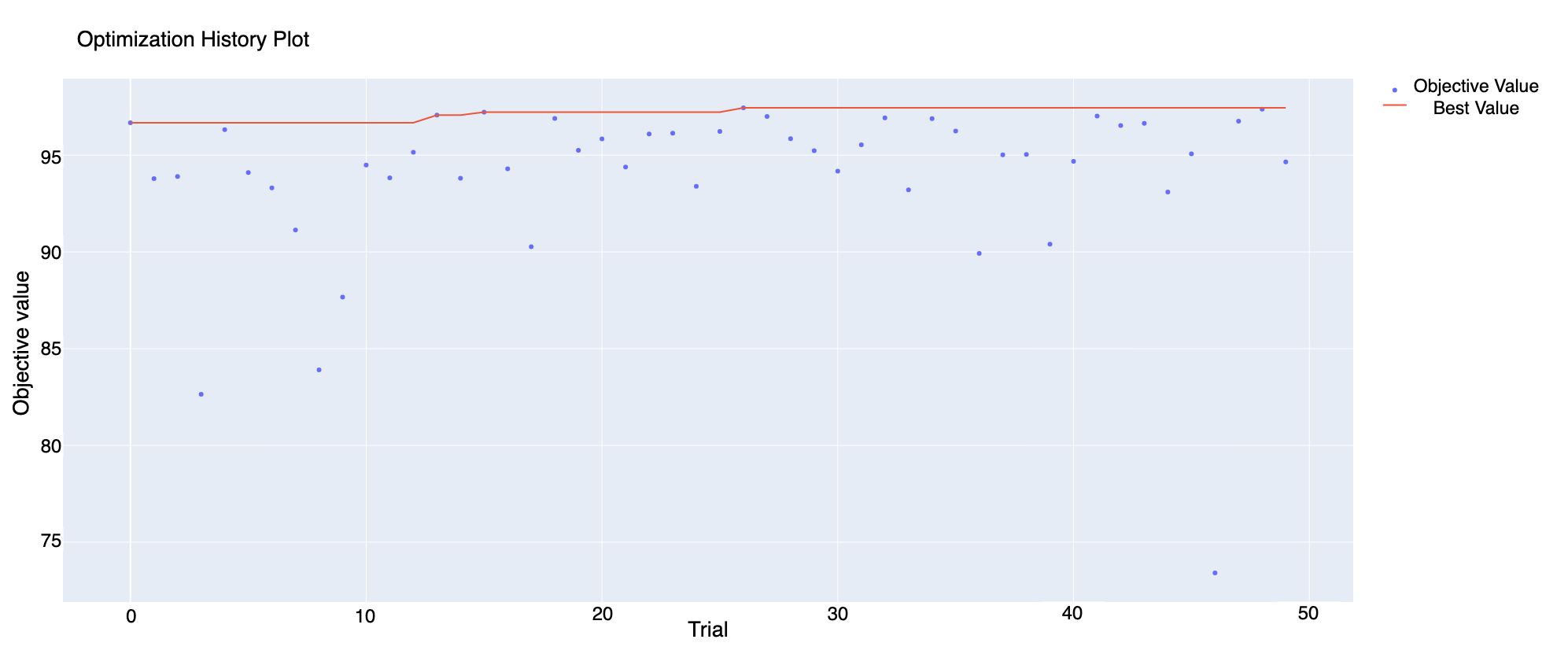}
\label{Optuna1}
\end{minipage}%
}%

\subfigure[Optuna contour plots]{
\begin{minipage}[t]{1\linewidth}
\centering
\includegraphics[height=3.8cm]{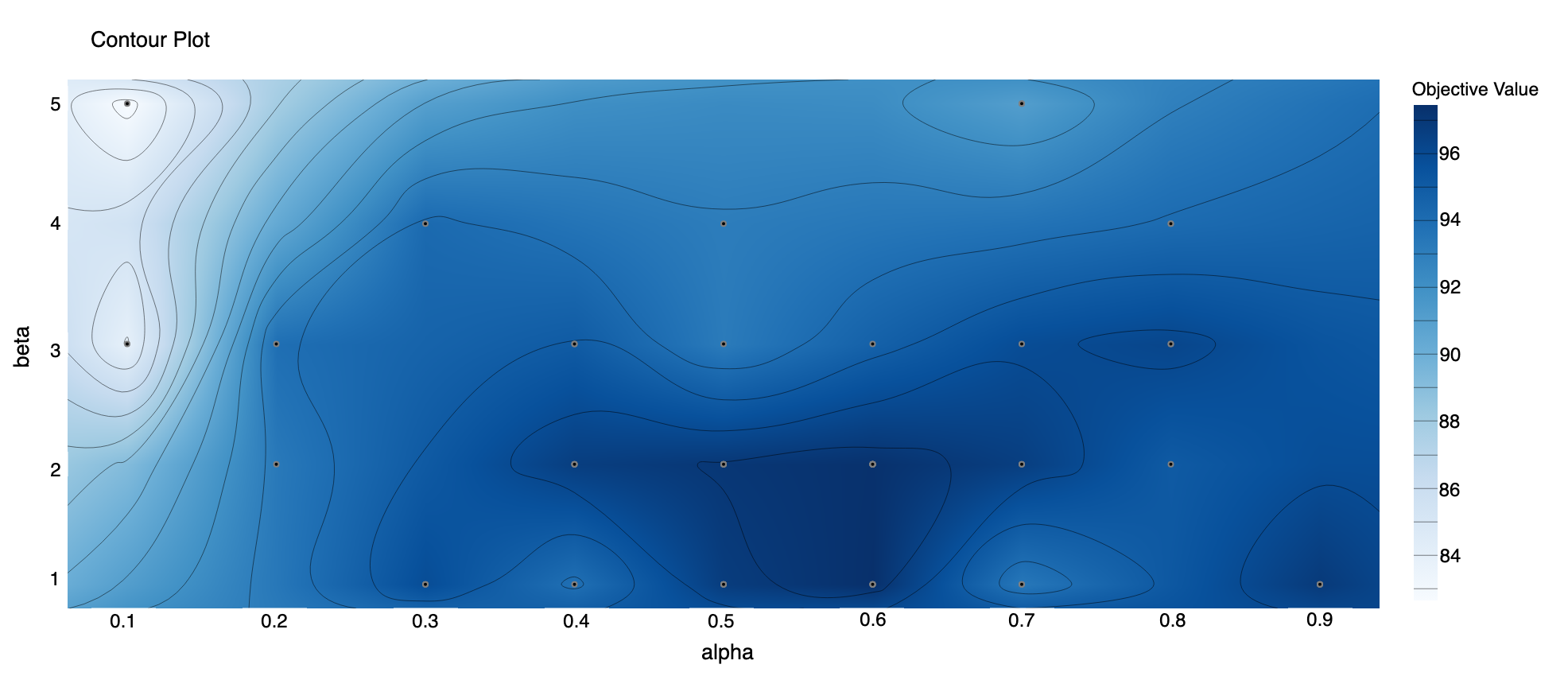}
\label{Optuna2}
\end{minipage}%
}%

\caption{Results of hyper-parameter optimization framework Optuna.}
\label{Optuna}
\end{figure}

In addition to using inefficient Manual Search hyper-parameter tunning, we explore an advanced hyper-parameter optimization framework, Optuna~\cite{akiba2019optuna} to find the optimal combination of $\alpha$ and $\beta$ automatically. Firstly, the range of parameters should be determined, with $\alpha \in [0.1,0.9]$ and $\beta \in [1,5]$. Optuna attempts to maximize the DER score and continuously learns from previous results to dynamically create a search space. From Fig.~\ref{alpha} it can be seen that BPP exhibits the highest sensitivity to hyper-parameters. Consequently, Fig.~\ref{Optuna} takes BPP as an example to illustrate the results. Fig.~\ref{Optuna1} plots the optimization history of Optuna. The horizontal axis represents the trial number,  with a total of 50 trials conducted in this experiment. The vertical axis represents the objective value, measured as the DER score. Each blue dot corresponds to the result of a trial, while the red curve represents the best value achieved from the initial trial up to the current one, showing a monotonically increasing trend. Fig.~\ref{Optuna2} displays contour plots illustrating the relationship between the two parameters and the objective value. The contour lines represent regions of equal objective values, with darker colors indicating higher values. The optimal objective value is achieved at $\beta=2$ and $\alpha \in (0.5, 0.6)$, aligning well with the results from manual parameter tuning shown in Fig.~\ref{alpha}.

Based on the results presented in Fig.~\ref{alpha} and Fig.~\ref{Optuna}, it can be observed that most of the optimal values are concentrated around $\beta=2$ and $\alpha \in (0,4,0.6)$. Finally, we set $\alpha$ to 0.6 and $\beta$ to 2, applying these settings to all experiments.

\begin{figure}[htbp]
\centerline

\subfigure[ASR]{
\begin{minipage}[t]{0.5\linewidth}
\centering
\includegraphics[height=3.5cm]{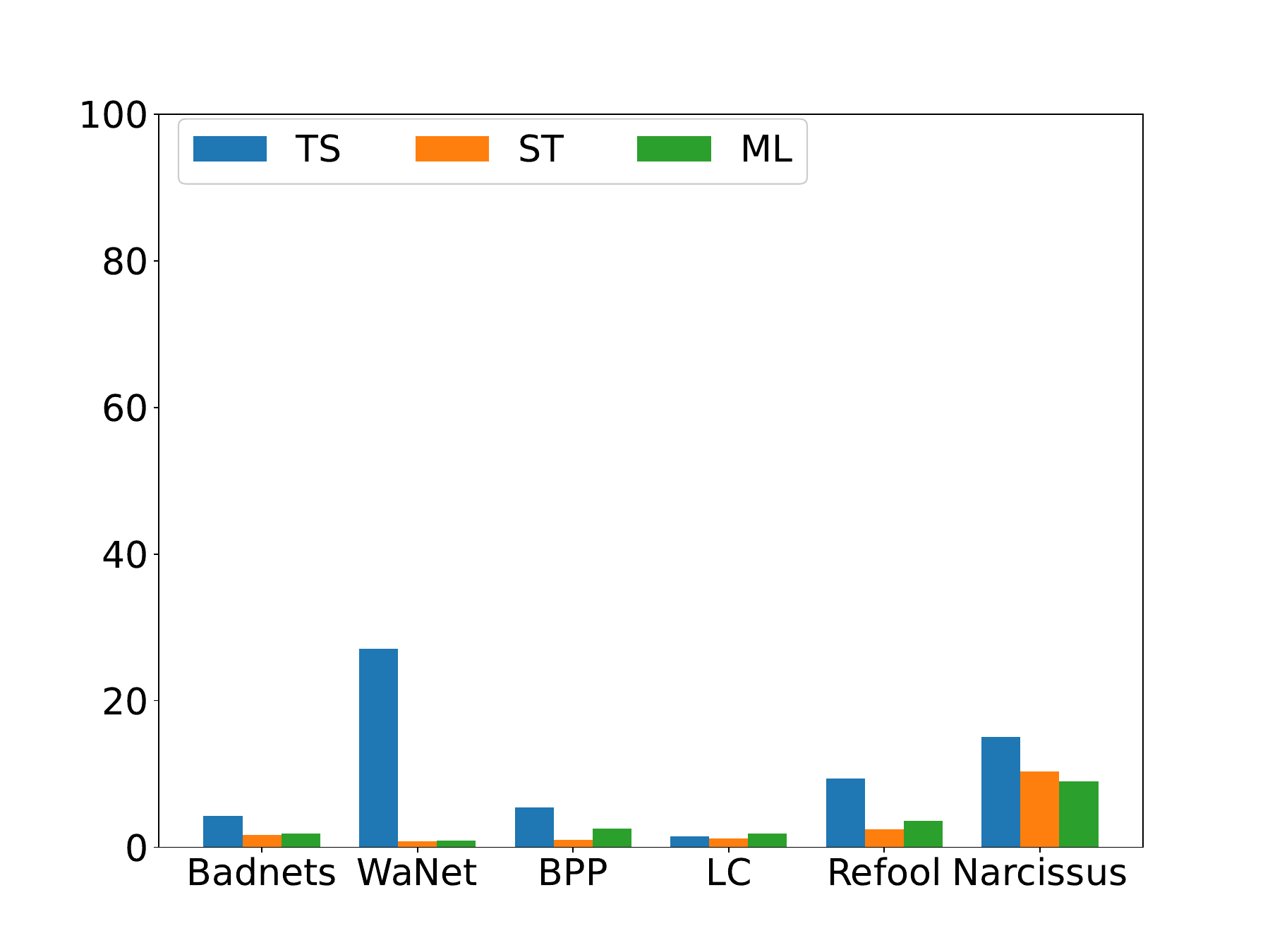}
\end{minipage}%
}%
\subfigure[BA]{
\begin{minipage}[t]{0.5\linewidth}
\centering
\includegraphics[height=3.5cm]{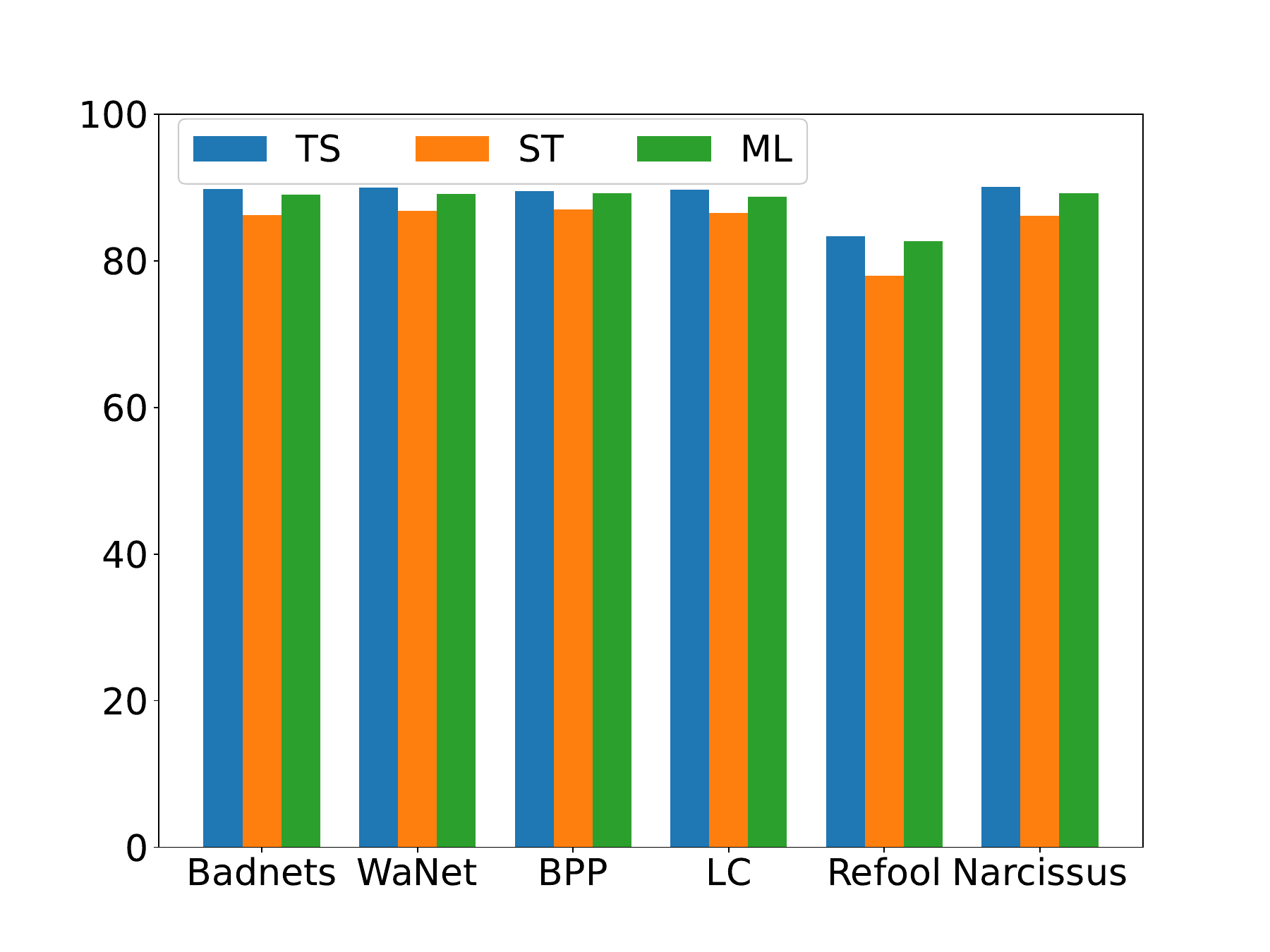}
\end{minipage}%
}%

\caption{Results of different teacher-student structures}
\label{teacher-student}
\end{figure}

\textbf{Effect of teacher-student structures} Considering the collaboration between two networks $F_{\theta_{tt}}$ and $F_{\theta_{nt}}$, there are three strategies to remove the backdoor: 1) $F_{\theta_{tt}}$ is the teacher and $F_{\theta_{nt}}$ is the student (TS), 2) $F_{\theta_{tt}}$ is the student and $F_{\theta_{nt}}$ is the teacher (ST), 3) $F_{\theta_{tt}}$ is the teacher and $F_{\theta_{nt}}$ is the student, but the teacher and student are updated simultaneously through mutual learning (ML). From Fig~\ref{teacher-student} it can be seen that BA and ASR of TS structure are both high. This is because the teacher model has higher accuracy but a backdoor has been injected. On the contrary, BA and ASR of the ST structure are both low, because the poisoned neurons of network $F_{\theta_{nt}}$ are alleviated in the NT process and the accuracy is also reduced at the same time. Therefore, whether it is the TS or  ST structure, the student network inherits the characteristics of the teacher network and is only good at one aspect. Through mutual learning, the strengths of the two networks can be combined, achieving the goal of reducing ASR while simultaneously maintaining a good BA.
 
\textbf{Effect of feature representations} We further investigate the defense performance of different feature representations. Table~\ref{fr} lists the results of two popular representations, attention maps (AM) and feature maps (FM). We apply the attention function $F_{sum}^{2} (F^{l})= {\textstyle \sum_{i=1}^C} \left | F_i^l  \right |^2$ to build attention maps, where $F^l\in R^{C_l\times H_l\times W_l}$ represents feature maps at the $l$-th layer. The authors in~\cite{zagoruyko2016paying} demonstrated that student models can significantly improve their performance by mimicking attention maps of teachers. However, in our case, an attention map does not yield better results for most attacks except Narcissus. We attribute this to the fact that in traditional knowledge distillation, teachers are much more powerful than students and therefore attention maps carry more compact and important information. But the difference between teacher $F_{\theta_{tt}}$ and student $F_{\theta_{nt}}$ is small, it is only reflected in the activation of backdoor neurons, as teacher and student apply the same structure and training data. Integrating the information of feature maps will further reduce the difference, resulting in more limited information that the student can learn. This could be the reason why transferring the original feature maps instead of attention maps is more effective.

\begin{table}[htbp]
	\centering
	\caption{Results of different feature represenattions.}
	\label{Defense results.}  
        \resizebox{\linewidth}{!}{
	\begin{tabular}{@{}l|l|l|l|l|l|l|l|l @{}}
	   \hline
            \multirow{2}{*}{FR}&~& \multicolumn{6}{c|}{Attack} &  \multirow{2}{*}{AVG}\\
            \cline{3-8}
             ~&~ & Badnets & WaNet & BPP & LC & Refool & Narcissus & ~ \\
            \hline
            \multirow{3}{*}{AM}&ASR& 1.91 & 0.92 & 2.90 & 2.08 & 4.44 & 8.71 & 3.49\\
            ~& PA& 87.12 & 86.81 & 67.83 & 84.35 & 42.66 & 75.34 & 74.02\\
            ~& BA & 88.54 & 89.01 & 89.27 & 88.58 & 83.36 & 89.14 & 87.98\\
            \hline
             \multirow{3}{*}{FM}&ASR& 1.84 & 0.89 & 2.57 & 1.92 & 3.62 & 8.96 & 3.30\\
            ~& PA& 87.72 & 86.96 & 67.40 & 84.46 & 43.67 & 75.25 & 74.24\\
            ~& BA& 89.06 & 89.15 & 89.24 & 88.78 & 82.71 & 89.20 & 88.02\\
            \hline
      
	\end{tabular}
 }
\vspace{-1em}
\label{fr}
\end{table}

\begin{figure*}[htbp]
\centerline

\subfigure[Clean Ratio 1\%]{
\begin{minipage}[t]{0.33\linewidth}
\centering
\includegraphics[height=6cm]{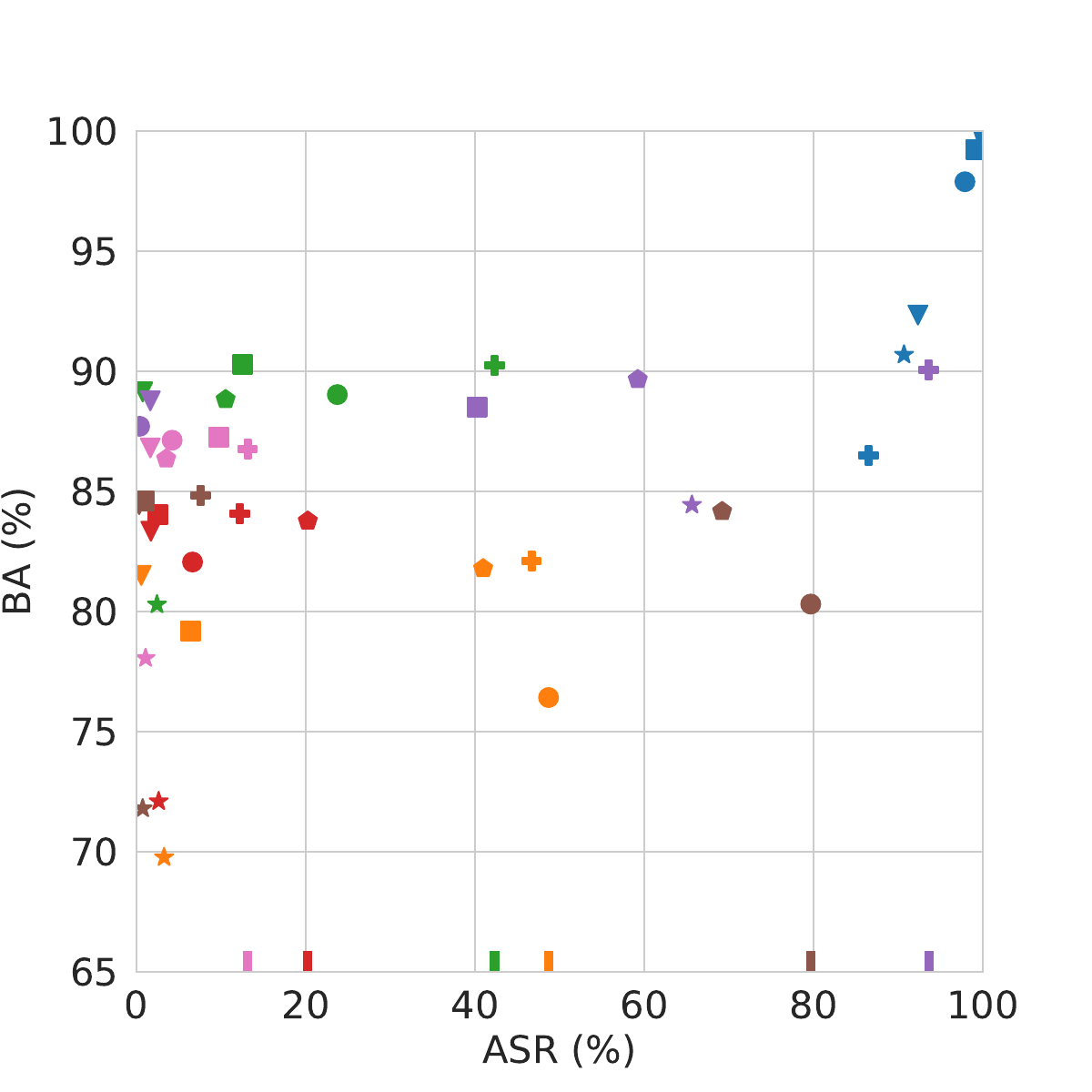}
\end{minipage}%
}%
\subfigure[Clean Ratio 5\%]{
\begin{minipage}[t]{0.33\linewidth}
\centering
\includegraphics[height=6cm]{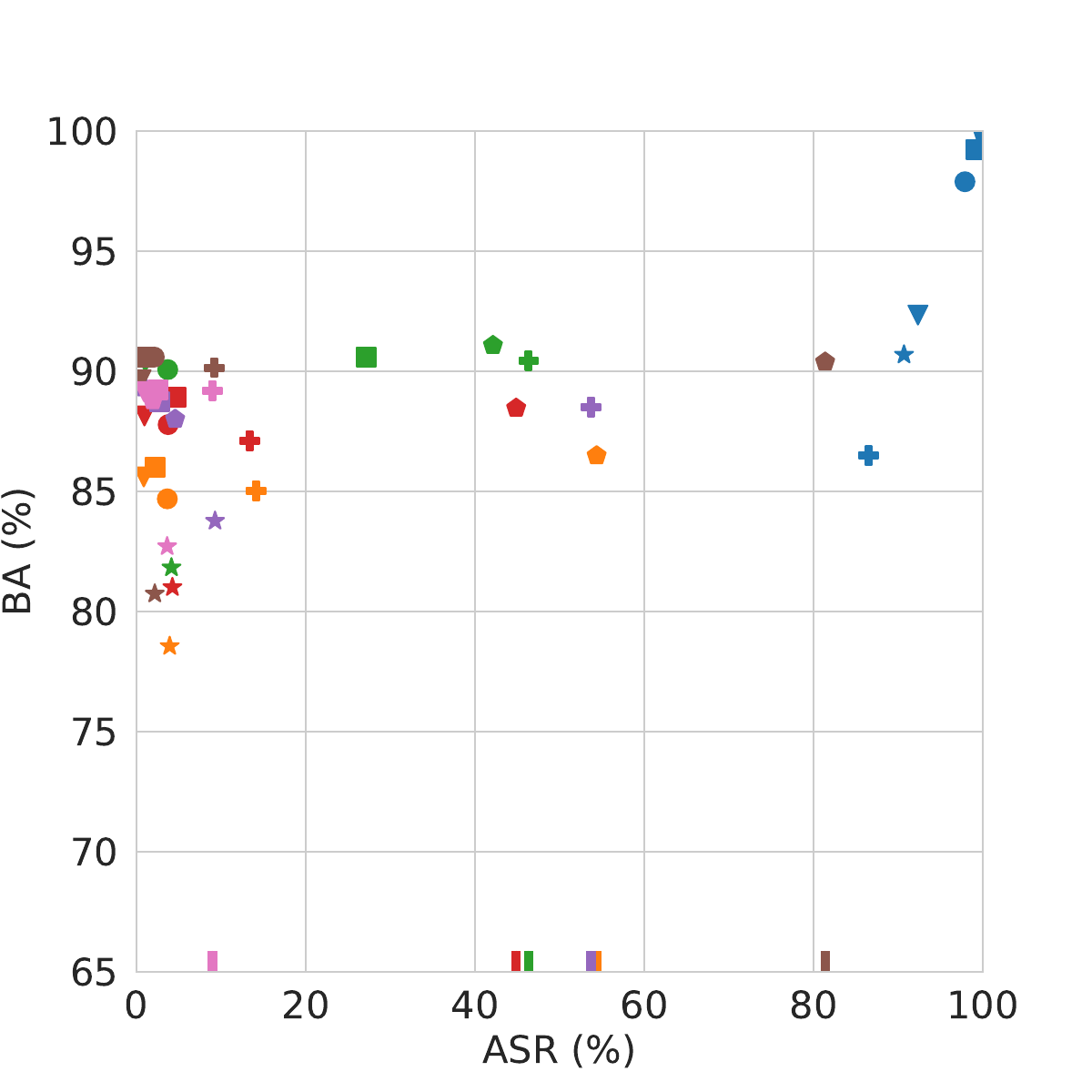}
\end{minipage}%
}%
\subfigure[Clean Ratio 10\%]{
\begin{minipage}[t]{0.33\linewidth}
\centering
\includegraphics[height=6cm]{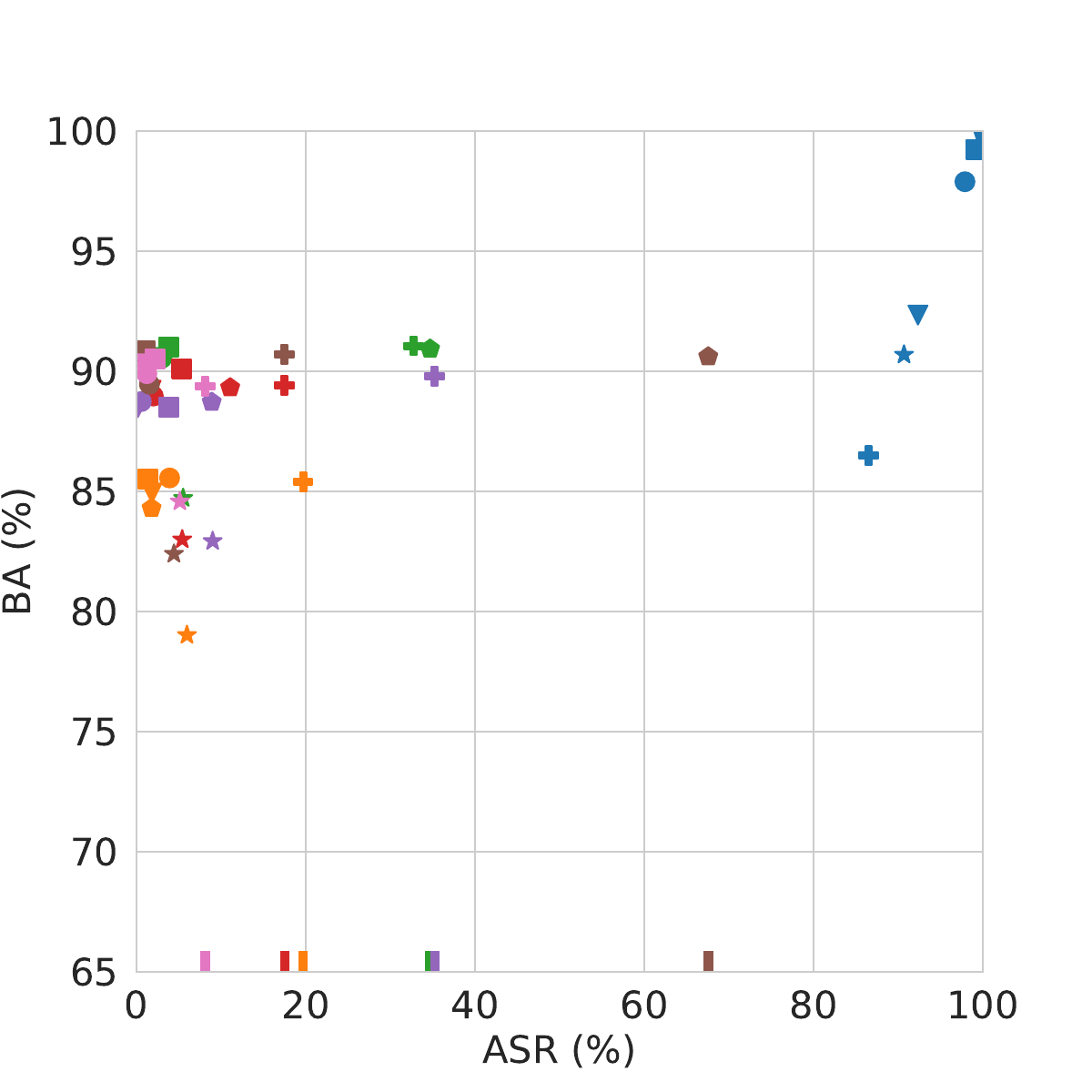}
\end{minipage}%
\label{ratio10}
}%

\subfigure{
\begin{minipage}[t]{\linewidth}
\centering
\includegraphics[height=0.5cm]{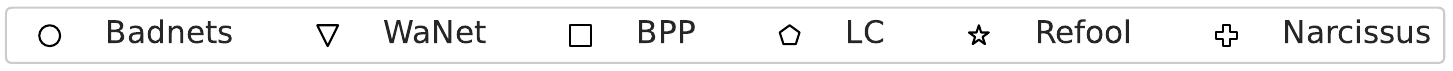}
\end{minipage}%
}%

\subfigure{
\begin{minipage}[t]{\linewidth}
\centering
\includegraphics[height=0.5cm]{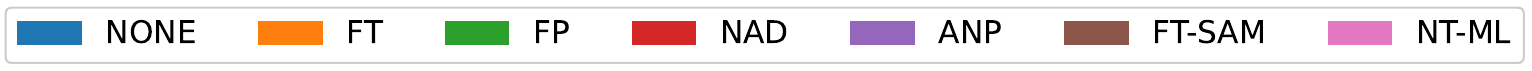}
\end{minipage}%
}%

\caption{Defense results under different size of clean data.}
\label{clean ratio}
\end{figure*}

\textbf{Effect of clean data size} The size of the local clean dataset also influences the defense performance. Here we compare the defenses using $1\%, 5\%,$ and $10\%$ of the training set, respectively. Fig.~\ref{clean ratio} shows the performance of attack-defense pairs, with patterns representing different attacks and colors representing different defenses. The vertical axis is BA, the horizontal axis is ASR, and the blue patterns show the results without defense. The successful attacker should achieve high BA and ASR simultaneously, meaning that the closer to the upper right corner, the better the attack. The objective for the defender is high BA and low ASR, reflected in the region in the upper left corner.

It can be observed that the performance improves with the increase of data size. When the clean set is only $1\%,$ several defenses such as ANP against Narcissus and FT-SAM against Badnets completely fail because their ASR almost exceeds 80\%. And there are also more points within the horizontal axis range of 20\% -80\% and the vertical axis range of 65\% -80\%. Compared to 1\% clean set, the BA of applying 5\% set is improved as there is only one point below 80\%. And when the clean set is expanded to 10\%, the defenses under the LC and Refool attacks are improved. ASR of most defenses has been successfully reduced to below $40\%$, as shown in Fig.~\ref{ratio10}. 

The results of NT-ML are displayed in pink patterns, indicating that it also performs well on a smaller clean dataset. A scenario considered in practical applications where the defender does not know what type of attack is used, we show the highest ASR for each defense on the x-axis. The worst defenses are mostly caused by clean-label attacks LC and Narcissus. It can be seen that the worst-case results of NT-ML are superior to other methods in all three clean ratios. In summary, NT-ML outperforms other strategies significantly on very small dataset ($1\%$), while the gap with other defense algorithms is narrowed on $5\%$ and $10\%$ clean set size.

\section{Conclusions}\label{section5}
 
 This paper provides the novel defense strategy {\em NT-ML} against backdoor attacks. We decouple the outputs of the standard trained model into a target class and a non-target class, and then utilize the non-target predictions, which are more informative, to improve the confidence of the correct category on poisoned inputs. We apply mutual learning and find that the teacher network (through standard training) can help the student network (through non-target training) to successfully remove the backdoor. Experiments on benchmark datasets indicate that NT-ML can defend very well against $6$ state-of-the-art backdoor attacks. It performs better than the other $4$ baseline defenses which we use for comparison.

Considering that some users do not have their own datasets, in the future we will investigate how to implement a defense without additional datasets. Secondly, we noticed that there is little work on successful distillation based defenses against attacks. Assuming the attacker knows that distillation will be used for the defense, an interesting research direction is how to destroy this defense system. This can be the starting point to constructing a more powerful defense against this new attack.

\section*{Acknowledgment}

This work is supported by the China Scholarship Council file Nr. 202008610227.

\bibliographystyle{IEEEtran}
\bibliography{sn-bibliography}



\end{document}